\documentclass{article}
\PassOptionsToPackage{numbers, compress}{natbib}
\usepackage[preprint]{neurips_2026}

\usepackage[utf8]{inputenc} 
\usepackage[T1]{fontenc}    
\usepackage{hyperref}       
\usepackage{url}            
\usepackage{booktabs}       
\usepackage{tabularx}       
\usepackage{amsfonts}       
\usepackage{nicefrac}       
\usepackage{microtype}      
\usepackage{xcolor}         

\usepackage{graphicx}
\usepackage{subfigure}
\usepackage{multirow}

\usepackage{amsmath}
\usepackage{amssymb}
\usepackage{mathtools}
\usepackage{amsthm}

\usepackage[capitalize,noabbrev]{cleveref}

\usepackage{tikz}
\usetikzlibrary{shapes.geometric, arrows.meta, positioning, decorations.pathreplacing, calc, fit, backgrounds}
\usepackage{enumitem} 
\usepackage[most]{tcolorbox} 

\theoremstyle{plain}

\theoremstyle{definition}

\theoremstyle{remark}

\usepackage[textsize=tiny]{todonotes}

\newcommand{\frameworkName}{PAVE}
\newcommand{\datasetName}{OpenRCA~2.0}

\definecolor{fillorange}{rgb}{0.88,0.45,0.05}
\definecolor{reviewred}{rgb}{0.78,0.10,0.10}

\definecolor{jsonstring}{HTML}{A31515}
\definecolor{jsonnumber}{HTML}{098658}
\definecolor{jsonbrace}{HTML}{0451A5}
\definecolor{jsonkey}{HTML}{0451A5}

\lstdefinelanguage{json}{
  basicstyle=\small\ttfamily,
  columns=fullflexible,
  keepspaces=true,
  showstringspaces=false,
  breaklines=true,
  literate=
    *{:}{{{\color{black}{:}}}}{1}
    {,}{{{\color{black}{,}}}}{1}
    {\{}{{{\color{jsonbrace}{\{}}}}{1}
    {\}}{{{\color{jsonbrace}{\}}}}}{1}
    {[}{{{\color{jsonbrace}{[}}}}{1}
    {]}{{{\color{jsonbrace}{]}}}}{1},
  string=[s]{"}{"},
  stringstyle=\color{jsonstring},
  morecomment=[l]{//},
  commentstyle=\color{gray}\itshape,
}

\newtcolorbox{systempromptbox}[1][]{%
  enhanced, breakable,
  colback=blue!3, colframe=blue!40!black,
  fonttitle=\bfseries\small\sffamily,
  title={#1},
  rounded corners, arc=3mm,
  left=4pt, right=4pt, top=2pt, bottom=2pt,
  boxrule=0.6pt,
  fontupper=\small\ttfamily,
  before upper={\parindent=0pt\parskip=0.4em\obeylines},
}
\newtcolorbox{userpromptbox}[1][]{%
  enhanced, breakable,
  colback=green!3, colframe=green!40!black,
  fonttitle=\bfseries\small\sffamily,
  title={#1},
  rounded corners, arc=3mm,
  left=4pt, right=4pt, top=2pt, bottom=2pt,
  boxrule=0.6pt,
  fontupper=\small\ttfamily,
  before upper={\parindent=0pt\parskip=0.4em\obeylines},
}
\newtcblisting{schemabox}[1][]{%
  enhanced, breakable,
  colback=gray!4, colframe=gray!60!black,
  fonttitle=\bfseries\small\sffamily,
  title={#1},
  rounded corners, arc=3mm,
  left=4pt, right=4pt, top=2pt, bottom=2pt,
  boxrule=0.6pt,
  listing only,
  listing options={
    language=json,
    basicstyle=\small\ttfamily,
    columns=fullflexible,
    keepspaces=true,
    xleftmargin=0pt,
  },
}

\newcommand{\totalRawInjections}{8137}

\newcommand{\faultTypeCount}{27}
\newcommand{\avgCausalEdges}{7.5}

\newcommand{\totalValidLite}{500}

\newcommand{\systemCount}{3}
\newcommand{\evalCountTt}{320}
\newcommand{\evalCountOtel}{38}
\newcommand{\evalCountHr}{142}
\newcommand{\faultKindsTt}{25}
\newcommand{\faultKindsOtel}{9}
\newcommand{\faultKindsHr}{5}
\newcommand{\meanLpTt}{3.0}
\newcommand{\meanLpOtel}{2.5}
\newcommand{\meanLpHr}{2.1}
\newcommand{\meanLp}{2.7}
\newcommand{\avgCausalEdgesTt}{10.1}
\newcommand{\avgCausalEdgesOtel}{3.0}
\newcommand{\avgCausalEdgesHr}{2.7}

\newcommand{\opusLiteTtAnySvc}{75.3}
\newcommand{\opusLiteTtPathReach}{55.9}
\newcommand{\opusLiteTtPathReachHit}{22.2}

\newcommand{\opusLiteTtEM}{15.6}
\newcommand{\opusLiteTtPrecision}{23.4}
\newcommand{\opusLiteTtRecall}{22.7}
\newcommand{\opusLiteTtF}{22.3}
\newcommand{\opusLiteTtNodeF}{58.9}
\newcommand{\opusLiteTtEdgeF}{36.3}

\newcommand{\opusLiteTtSqlExec}{98.7}

\newcommand{\opusLiteOtelAnySvc}{81.6}
\newcommand{\opusLiteOtelPathReach}{81.6}
\newcommand{\opusLiteOtelPathReachHit}{60.5}

\newcommand{\opusLiteOtelEM}{26.3}
\newcommand{\opusLiteOtelPrecision}{51.8}
\newcommand{\opusLiteOtelRecall}{52.6}
\newcommand{\opusLiteOtelF}{50.5}
\newcommand{\opusLiteOtelNodeF}{77.4}
\newcommand{\opusLiteOtelEdgeF}{66.5}

\newcommand{\opusLiteOtelSqlExec}{98.7}

\newcommand{\opusLiteHrAnySvc}{99.3}
\newcommand{\opusLiteHrPathReach}{98.6}
\newcommand{\opusLiteHrPathReachHit}{94.4}

\newcommand{\opusLiteHrEM}{54.9}
\newcommand{\opusLiteHrPrecision}{94.4}
\newcommand{\opusLiteHrRecall}{76.1}
\newcommand{\opusLiteHrF}{81.9}
\newcommand{\opusLiteHrNodeF}{84.2}
\newcommand{\opusLiteHrEdgeF}{74.7}

\newcommand{\opusLiteHrSqlExec}{94.4}

\newcommand{\opusLiteAnySvc}{82.6}
\newcommand{\opusLitePathReach}{70.0}
\newcommand{\opusLitePathReachHit}{45.6}

\newcommand{\opusLiteEM}{27.6}
\newcommand{\opusLitePrecision}{45.7}
\newcommand{\opusLiteRecall}{40.1}
\newcommand{\opusLiteF}{41.4}
\newcommand{\opusLiteNodeF}{67.5}
\newcommand{\opusLiteEdgeF}{49.5}

\newcommand{\opusLiteSqlExec}{97.5}

\newcommand{\sonnetLiteTtAnySvc}{70.3}
\newcommand{\sonnetLiteTtPathReach}{30.0}
\newcommand{\sonnetLiteTtPathReachHit}{16.9}

\newcommand{\sonnetLiteTtEM}{15.0}
\newcommand{\sonnetLiteTtPrecision}{21.9}
\newcommand{\sonnetLiteTtRecall}{20.9}
\newcommand{\sonnetLiteTtF}{20.8}
\newcommand{\sonnetLiteTtNodeF}{55.1}
\newcommand{\sonnetLiteTtEdgeF}{34.1}

\newcommand{\sonnetLiteTtSqlExec}{96.8}

\newcommand{\sonnetLiteOtelAnySvc}{89.5}
\newcommand{\sonnetLiteOtelPathReach}{81.6}
\newcommand{\sonnetLiteOtelPathReachHit}{57.9}

\newcommand{\sonnetLiteOtelEM}{34.2}
\newcommand{\sonnetLiteOtelPrecision}{61.4}
\newcommand{\sonnetLiteOtelRecall}{57.9}
\newcommand{\sonnetLiteOtelF}{57.6}
\newcommand{\sonnetLiteOtelNodeF}{72.5}
\newcommand{\sonnetLiteOtelEdgeF}{62.1}

\newcommand{\sonnetLiteOtelSqlExec}{96.6}

\newcommand{\sonnetLiteHrAnySvc}{97.2}
\newcommand{\sonnetLiteHrPathReach}{95.8}
\newcommand{\sonnetLiteHrPathReachHit}{94.4}

\newcommand{\sonnetLiteHrEM}{49.3}
\newcommand{\sonnetLiteHrPrecision}{95.1}
\newcommand{\sonnetLiteHrRecall}{72.9}
\newcommand{\sonnetLiteHrF}{80.2}
\newcommand{\sonnetLiteHrNodeF}{83.7}
\newcommand{\sonnetLiteHrEdgeF}{75.4}

\newcommand{\sonnetLiteHrSqlExec}{92.8}

\newcommand{\sonnetLiteAnySvc}{79.4}
\newcommand{\sonnetLitePathReach}{52.6}
\newcommand{\sonnetLitePathReachHit}{42.0}

\newcommand{\sonnetLiteEM}{26.2}
\newcommand{\sonnetLitePrecision}{45.7}
\newcommand{\sonnetLiteRecall}{38.5}
\newcommand{\sonnetLiteF}{40.5}
\newcommand{\sonnetLiteNodeF}{64.5}
\newcommand{\sonnetLiteEdgeF}{48.0}

\newcommand{\sonnetLiteSqlExec}{95.6}

\newcommand{\geminiLiteTtAnySvc}{75.0}
\newcommand{\geminiLiteTtPathReach}{57.8}
\newcommand{\geminiLiteTtPathReachHit}{24.4}

\newcommand{\geminiLiteTtEM}{18.1}
\newcommand{\geminiLiteTtPrecision}{28.0}
\newcommand{\geminiLiteTtRecall}{24.1}
\newcommand{\geminiLiteTtF}{25.2}
\newcommand{\geminiLiteTtNodeF}{59.8}
\newcommand{\geminiLiteTtEdgeF}{27.1}

\newcommand{\geminiLiteTtSqlExec}{98.7}

\newcommand{\geminiLiteOtelAnySvc}{81.6}
\newcommand{\geminiLiteOtelPathReach}{76.3}
\newcommand{\geminiLiteOtelPathReachHit}{60.5}

\newcommand{\geminiLiteOtelEM}{36.8}
\newcommand{\geminiLiteOtelPrecision}{67.1}
\newcommand{\geminiLiteOtelRecall}{52.6}
\newcommand{\geminiLiteOtelF}{57.5}
\newcommand{\geminiLiteOtelNodeF}{77.3}
\newcommand{\geminiLiteOtelEdgeF}{31.9}

\newcommand{\geminiLiteOtelSqlExec}{99.0}

\newcommand{\geminiLiteHrAnySvc}{97.9}
\newcommand{\geminiLiteHrPathReach}{97.2}
\newcommand{\geminiLiteHrPathReachHit}{97.2}

\newcommand{\geminiLiteHrEM}{52.8}
\newcommand{\geminiLiteHrPrecision}{96.5}
\newcommand{\geminiLiteHrRecall}{75.0}
\newcommand{\geminiLiteHrF}{82.2}
\newcommand{\geminiLiteHrNodeF}{83.8}
\newcommand{\geminiLiteHrEdgeF}{60.0}

\newcommand{\geminiLiteHrSqlExec}{98.9}

\newcommand{\geminiLiteAnySvc}{82.0}
\newcommand{\geminiLitePathReach}{70.4}
\newcommand{\geminiLitePathReachHit}{47.8}

\newcommand{\geminiLiteEM}{29.4}
\newcommand{\geminiLitePrecision}{50.4}
\newcommand{\geminiLiteRecall}{40.7}
\newcommand{\geminiLiteF}{43.8}
\newcommand{\geminiLiteNodeF}{67.9}
\newcommand{\geminiLiteEdgeF}{36.8}

\newcommand{\geminiLiteSqlExec}{98.8}

\newcommand{\mimoLiteTtAnySvc}{68.8}
\newcommand{\mimoLiteTtPathReach}{42.2}
\newcommand{\mimoLiteTtPathReachHit}{16.6}

\newcommand{\mimoLiteTtEM}{11.2}
\newcommand{\mimoLiteTtPrecision}{17.0}
\newcommand{\mimoLiteTtRecall}{18.0}
\newcommand{\mimoLiteTtF}{16.8}
\newcommand{\mimoLiteTtNodeF}{56.0}
\newcommand{\mimoLiteTtEdgeF}{32.1}

\newcommand{\mimoLiteTtSqlExec}{93.5}

\newcommand{\mimoLiteOtelAnySvc}{86.8}
\newcommand{\mimoLiteOtelPathReach}{76.3}
\newcommand{\mimoLiteOtelPathReachHit}{44.7}

\newcommand{\mimoLiteOtelEM}{31.6}
\newcommand{\mimoLiteOtelPrecision}{50.0}
\newcommand{\mimoLiteOtelRecall}{47.4}
\newcommand{\mimoLiteOtelF}{47.5}
\newcommand{\mimoLiteOtelNodeF}{71.7}
\newcommand{\mimoLiteOtelEdgeF}{56.0}

\newcommand{\mimoLiteOtelSqlExec}{94.4}

\newcommand{\mimoLiteHrAnySvc}{95.8}
\newcommand{\mimoLiteHrPathReach}{94.4}
\newcommand{\mimoLiteHrPathReachHit}{90.8}

\newcommand{\mimoLiteHrEM}{46.5}
\newcommand{\mimoLiteHrPrecision}{89.4}
\newcommand{\mimoLiteHrRecall}{69.4}
\newcommand{\mimoLiteHrF}{75.8}
\newcommand{\mimoLiteHrNodeF}{81.0}
\newcommand{\mimoLiteHrEdgeF}{70.5}

\newcommand{\mimoLiteHrSqlExec}{88.1}

\newcommand{\mimoLiteAnySvc}{77.8}
\newcommand{\mimoLitePathReach}{59.6}
\newcommand{\mimoLitePathReachHit}{39.8}

\newcommand{\mimoLiteEM}{22.8}
\newcommand{\mimoLitePrecision}{40.1}
\newcommand{\mimoLiteRecall}{34.8}
\newcommand{\mimoLiteF}{35.9}
\newcommand{\mimoLiteNodeF}{64.3}
\newcommand{\mimoLiteEdgeF}{44.9}

\newcommand{\mimoLiteSqlExec}{92.0}

\newcommand{\glmLiteTtAnySvc}{70.9}
\newcommand{\glmLiteTtPathReach}{56.9}
\newcommand{\glmLiteTtPathReachHit}{16.2}

\newcommand{\glmLiteTtEM}{9.4}
\newcommand{\glmLiteTtPrecision}{15.5}
\newcommand{\glmLiteTtRecall}{15.0}
\newcommand{\glmLiteTtF}{14.7}
\newcommand{\glmLiteTtNodeF}{59.8}
\newcommand{\glmLiteTtEdgeF}{38.6}

\newcommand{\glmLiteTtSqlExec}{93.9}

\newcommand{\glmLiteOtelAnySvc}{84.2}
\newcommand{\glmLiteOtelPathReach}{78.9}
\newcommand{\glmLiteOtelPathReachHit}{44.7}

\newcommand{\glmLiteOtelEM}{21.1}
\newcommand{\glmLiteOtelPrecision}{42.1}
\newcommand{\glmLiteOtelRecall}{39.5}
\newcommand{\glmLiteOtelF}{39.6}
\newcommand{\glmLiteOtelNodeF}{77.1}
\newcommand{\glmLiteOtelEdgeF}{68.7}

\newcommand{\glmLiteOtelSqlExec}{94.3}

\newcommand{\glmLiteHrAnySvc}{95.8}
\newcommand{\glmLiteHrPathReach}{95.8}
\newcommand{\glmLiteHrPathReachHit}{89.4}

\newcommand{\glmLiteHrEM}{49.3}
\newcommand{\glmLiteHrPrecision}{88.7}
\newcommand{\glmLiteHrRecall}{69.7}
\newcommand{\glmLiteHrF}{75.9}
\newcommand{\glmLiteHrNodeF}{80.3}
\newcommand{\glmLiteHrEdgeF}{71.9}

\newcommand{\glmLiteHrSqlExec}{92.2}

\newcommand{\glmLiteAnySvc}{79.0}
\newcommand{\glmLitePathReach}{69.6}
\newcommand{\glmLitePathReachHit}{39.2}

\newcommand{\glmLiteEM}{21.6}
\newcommand{\glmLitePrecision}{38.3}
\newcommand{\glmLiteRecall}{32.4}
\newcommand{\glmLiteF}{34.0}
\newcommand{\glmLiteNodeF}{66.9}
\newcommand{\glmLiteEdgeF}{50.4}

\newcommand{\glmLiteSqlExec}{93.4}

\newcommand{\kimiLiteTtAnySvc}{61.6}
\newcommand{\kimiLiteTtPathReach}{50.0}
\newcommand{\kimiLiteTtPathReachHit}{15.3}

\newcommand{\kimiLiteTtEM}{10.6}
\newcommand{\kimiLiteTtPrecision}{15.7}
\newcommand{\kimiLiteTtRecall}{15.9}
\newcommand{\kimiLiteTtF}{15.3}
\newcommand{\kimiLiteTtNodeF}{49.3}
\newcommand{\kimiLiteTtEdgeF}{30.8}

\newcommand{\kimiLiteTtSqlExec}{78.8}

\newcommand{\kimiLiteOtelAnySvc}{68.4}
\newcommand{\kimiLiteOtelPathReach}{65.8}
\newcommand{\kimiLiteOtelPathReachHit}{39.5}

\newcommand{\kimiLiteOtelEM}{18.4}
\newcommand{\kimiLiteOtelPrecision}{43.9}
\newcommand{\kimiLiteOtelRecall}{44.7}
\newcommand{\kimiLiteOtelF}{41.9}
\newcommand{\kimiLiteOtelNodeF}{61.0}
\newcommand{\kimiLiteOtelEdgeF}{51.5}

\newcommand{\kimiLiteOtelSqlExec}{72.9}

\newcommand{\kimiLiteHrAnySvc}{92.3}
\newcommand{\kimiLiteHrPathReach}{92.3}
\newcommand{\kimiLiteHrPathReachHit}{88.7}

\newcommand{\kimiLiteHrEM}{45.1}
\newcommand{\kimiLiteHrPrecision}{88.0}
\newcommand{\kimiLiteHrRecall}{66.9}
\newcommand{\kimiLiteHrF}{73.9}
\newcommand{\kimiLiteHrNodeF}{79.6}
\newcommand{\kimiLiteHrEdgeF}{72.1}

\newcommand{\kimiLiteHrSqlExec}{93.0}

\newcommand{\kimiLiteAnySvc}{70.8}
\newcommand{\kimiLitePathReach}{63.2}
\newcommand{\kimiLitePathReachHit}{38.0}

\newcommand{\kimiLiteEM}{21.0}
\newcommand{\kimiLitePrecision}{38.4}
\newcommand{\kimiLiteRecall}{32.6}
\newcommand{\kimiLiteF}{34.0}
\newcommand{\kimiLiteNodeF}{58.8}
\newcommand{\kimiLiteEdgeF}{44.1}

\newcommand{\kimiLiteSqlExec}{82.4}

\newcommand{\dsvfourLiteTtAnySvc}{54.1}
\newcommand{\dsvfourLiteTtPathReach}{35.6}
\newcommand{\dsvfourLiteTtPathReachHit}{13.4}

\newcommand{\dsvfourLiteTtEM}{10.9}
\newcommand{\dsvfourLiteTtPrecision}{15.0}
\newcommand{\dsvfourLiteTtRecall}{13.9}
\newcommand{\dsvfourLiteTtF}{14.1}
\newcommand{\dsvfourLiteTtNodeF}{42.1}
\newcommand{\dsvfourLiteTtEdgeF}{25.8}

\newcommand{\dsvfourLiteTtSqlExec}{70.1}

\newcommand{\dsvfourLiteOtelAnySvc}{65.8}
\newcommand{\dsvfourLiteOtelPathReach}{57.9}
\newcommand{\dsvfourLiteOtelPathReachHit}{39.5}

\newcommand{\dsvfourLiteOtelEM}{23.7}
\newcommand{\dsvfourLiteOtelPrecision}{42.5}
\newcommand{\dsvfourLiteOtelRecall}{38.2}
\newcommand{\dsvfourLiteOtelF}{39.4}
\newcommand{\dsvfourLiteOtelNodeF}{58.0}
\newcommand{\dsvfourLiteOtelEdgeF}{46.5}

\newcommand{\dsvfourLiteOtelSqlExec}{68.7}

\newcommand{\dsvfourLiteHrAnySvc}{91.5}
\newcommand{\dsvfourLiteHrPathReach}{90.8}
\newcommand{\dsvfourLiteHrPathReachHit}{85.2}

\newcommand{\dsvfourLiteHrEM}{40.8}
\newcommand{\dsvfourLiteHrPrecision}{84.9}
\newcommand{\dsvfourLiteHrRecall}{63.0}
\newcommand{\dsvfourLiteHrF}{70.3}
\newcommand{\dsvfourLiteHrNodeF}{78.1}
\newcommand{\dsvfourLiteHrEdgeF}{70.5}

\newcommand{\dsvfourLiteHrSqlExec}{87.4}

\newcommand{\dsvfourLiteAnySvc}{65.6}
\newcommand{\dsvfourLitePathReach}{53.0}
\newcommand{\dsvfourLitePathReachHit}{35.8}

\newcommand{\dsvfourLiteEM}{20.4}
\newcommand{\dsvfourLitePrecision}{36.9}
\newcommand{\dsvfourLiteRecall}{29.7}
\newcommand{\dsvfourLiteF}{32.0}
\newcommand{\dsvfourLiteNodeF}{53.5}
\newcommand{\dsvfourLiteEdgeF}{40.1}

\newcommand{\dsvfourLiteSqlExec}{74.9}

\newcommand{\qwenLiteTtAnySvc}{70.6}
\newcommand{\qwenLiteTtPathReach}{60.3}
\newcommand{\qwenLiteTtPathReachHit}{20.6}

\newcommand{\qwenLiteTtEM}{9.4}
\newcommand{\qwenLiteTtPrecision}{17.3}
\newcommand{\qwenLiteTtRecall}{19.1}
\newcommand{\qwenLiteTtF}{17.2}
\newcommand{\qwenLiteTtNodeF}{56.1}
\newcommand{\qwenLiteTtEdgeF}{36.7}

\newcommand{\qwenLiteTtSqlExec}{93.8}

\newcommand{\qwenLiteOtelAnySvc}{86.8}
\newcommand{\qwenLiteOtelPathReach}{76.3}
\newcommand{\qwenLiteOtelPathReachHit}{60.5}

\newcommand{\qwenLiteOtelEM}{21.1}
\newcommand{\qwenLiteOtelPrecision}{55.3}
\newcommand{\qwenLiteOtelRecall}{44.7}
\newcommand{\qwenLiteOtelF}{47.4}
\newcommand{\qwenLiteOtelNodeF}{77.1}
\newcommand{\qwenLiteOtelEdgeF}{67.5}

\newcommand{\qwenLiteOtelSqlExec}{89.4}

\newcommand{\qwenLiteHrAnySvc}{96.5}
\newcommand{\qwenLiteHrPathReach}{96.5}
\newcommand{\qwenLiteHrPathReachHit}{91.5}

\newcommand{\qwenLiteHrEM}{40.8}
\newcommand{\qwenLiteHrPrecision}{88.8}
\newcommand{\qwenLiteHrRecall}{68.0}
\newcommand{\qwenLiteHrF}{74.4}
\newcommand{\qwenLiteHrNodeF}{82.0}
\newcommand{\qwenLiteHrEdgeF}{69.8}

\newcommand{\qwenLiteHrSqlExec}{92.0}

\newcommand{\qwenLiteAnySvc}{79.2}
\newcommand{\qwenLitePathReach}{71.8}
\newcommand{\qwenLitePathReachHit}{43.8}

\newcommand{\qwenLiteEM}{19.2}
\newcommand{\qwenLitePrecision}{40.5}
\newcommand{\qwenLiteRecall}{34.9}
\newcommand{\qwenLiteF}{35.7}
\newcommand{\qwenLiteNodeF}{65.1}
\newcommand{\qwenLiteEdgeF}{48.5}

\newcommand{\qwenLiteSqlExec}{92.9}

\newcommand{\hyLiteTtAnySvc}{60.6}
\newcommand{\hyLiteTtPathReach}{34.7}
\newcommand{\hyLiteTtPathReachHit}{12.2}

\newcommand{\hyLiteTtEM}{6.2}
\newcommand{\hyLiteTtPrecision}{11.0}
\newcommand{\hyLiteTtRecall}{14.4}
\newcommand{\hyLiteTtF}{11.7}
\newcommand{\hyLiteTtNodeF}{43.2}
\newcommand{\hyLiteTtEdgeF}{22.9}

\newcommand{\hyLiteTtSqlExec}{89.5}

\newcommand{\hyLiteOtelAnySvc}{63.2}
\newcommand{\hyLiteOtelPathReach}{57.9}
\newcommand{\hyLiteOtelPathReachHit}{39.5}

\newcommand{\hyLiteOtelEM}{21.1}
\newcommand{\hyLiteOtelPrecision}{42.1}
\newcommand{\hyLiteOtelRecall}{36.8}
\newcommand{\hyLiteOtelF}{37.7}
\newcommand{\hyLiteOtelNodeF}{69.9}
\newcommand{\hyLiteOtelEdgeF}{50.3}

\newcommand{\hyLiteOtelSqlExec}{98.0}

\newcommand{\hyLiteHrAnySvc}{96.5}
\newcommand{\hyLiteHrPathReach}{94.4}
\newcommand{\hyLiteHrPathReachHit}{83.8}

\newcommand{\hyLiteHrEM}{41.5}
\newcommand{\hyLiteHrPrecision}{81.5}
\newcommand{\hyLiteHrRecall}{64.1}
\newcommand{\hyLiteHrF}{69.4}
\newcommand{\hyLiteHrNodeF}{80.6}
\newcommand{\hyLiteHrEdgeF}{69.8}

\newcommand{\hyLiteHrSqlExec}{96.6}

\newcommand{\hyLiteAnySvc}{71.0}
\newcommand{\hyLitePathReach}{53.4}
\newcommand{\hyLitePathReachHit}{34.6}

\newcommand{\hyLiteEM}{17.4}
\newcommand{\hyLitePrecision}{33.4}
\newcommand{\hyLiteRecall}{30.2}
\newcommand{\hyLiteF}{30.1}
\newcommand{\hyLiteNodeF}{55.9}
\newcommand{\hyLiteEdgeF}{38.3}

\newcommand{\hyLiteSqlExec}{92.1}

\newcommand{\seedLiteTtAnySvc}{72.5}
\newcommand{\seedLiteTtPathReach}{53.8}
\newcommand{\seedLiteTtPathReachHit}{16.2}

\newcommand{\seedLiteTtEM}{8.8}
\newcommand{\seedLiteTtPrecision}{14.1}
\newcommand{\seedLiteTtRecall}{16.1}
\newcommand{\seedLiteTtF}{14.3}
\newcommand{\seedLiteTtNodeF}{57.3}
\newcommand{\seedLiteTtEdgeF}{35.7}

\newcommand{\seedLiteTtSqlExec}{89.8}

\newcommand{\seedLiteOtelAnySvc}{73.7}
\newcommand{\seedLiteOtelPathReach}{71.1}
\newcommand{\seedLiteOtelPathReachHit}{34.2}

\newcommand{\seedLiteOtelEM}{10.5}
\newcommand{\seedLiteOtelPrecision}{26.6}
\newcommand{\seedLiteOtelRecall}{25.0}
\newcommand{\seedLiteOtelF}{24.2}
\newcommand{\seedLiteOtelNodeF}{72.9}
\newcommand{\seedLiteOtelEdgeF}{63.3}

\newcommand{\seedLiteOtelSqlExec}{87.2}

\newcommand{\seedLiteHrAnySvc}{93.7}
\newcommand{\seedLiteHrPathReach}{93.7}
\newcommand{\seedLiteHrPathReachHit}{70.4}

\newcommand{\seedLiteHrEM}{29.6}
\newcommand{\seedLiteHrPrecision}{67.1}
\newcommand{\seedLiteHrRecall}{50.7}
\newcommand{\seedLiteHrF}{55.9}
\newcommand{\seedLiteHrNodeF}{78.7}
\newcommand{\seedLiteHrEdgeF}{70.5}

\newcommand{\seedLiteHrSqlExec}{93.3}

\newcommand{\seedLiteAnySvc}{78.6}
\newcommand{\seedLitePathReach}{66.4}
\newcommand{\seedLitePathReachHit}{33.0}

\newcommand{\seedLiteEM}{14.8}
\newcommand{\seedLitePrecision}{30.1}
\newcommand{\seedLiteRecall}{26.6}
\newcommand{\seedLiteF}{26.8}
\newcommand{\seedLiteNodeF}{64.5}
\newcommand{\seedLiteEdgeF}{47.7}

\newcommand{\seedLiteSqlExec}{90.6}

\newcommand{\minimaxLiteTtAnySvc}{64.7}
\newcommand{\minimaxLiteTtPathReach}{33.1}
\newcommand{\minimaxLiteTtPathReachHit}{7.8}

\newcommand{\minimaxLiteTtEM}{1.9}
\newcommand{\minimaxLiteTtPrecision}{7.1}
\newcommand{\minimaxLiteTtRecall}{11.4}
\newcommand{\minimaxLiteTtF}{8.3}
\newcommand{\minimaxLiteTtNodeF}{46.9}
\newcommand{\minimaxLiteTtEdgeF}{21.2}

\newcommand{\minimaxLiteTtSqlExec}{86.6}

\newcommand{\minimaxLiteOtelAnySvc}{52.6}
\newcommand{\minimaxLiteOtelPathReach}{39.5}
\newcommand{\minimaxLiteOtelPathReachHit}{15.8}

\newcommand{\minimaxLiteOtelEM}{2.6}
\newcommand{\minimaxLiteOtelPrecision}{14.5}
\newcommand{\minimaxLiteOtelRecall}{18.4}
\newcommand{\minimaxLiteOtelF}{15.4}
\newcommand{\minimaxLiteOtelNodeF}{62.2}
\newcommand{\minimaxLiteOtelEdgeF}{43.6}

\newcommand{\minimaxLiteOtelSqlExec}{88.3}

\newcommand{\minimaxLiteHrAnySvc}{85.9}
\newcommand{\minimaxLiteHrPathReach}{78.2}
\newcommand{\minimaxLiteHrPathReachHit}{70.4}

\newcommand{\minimaxLiteHrEM}{19.0}
\newcommand{\minimaxLiteHrPrecision}{49.5}
\newcommand{\minimaxLiteHrRecall}{54.6}
\newcommand{\minimaxLiteHrF}{49.7}
\newcommand{\minimaxLiteHrNodeF}{72.5}
\newcommand{\minimaxLiteHrEdgeF}{43.4}

\newcommand{\minimaxLiteHrSqlExec}{87.3}

\newcommand{\minimaxLiteAnySvc}{69.8}
\newcommand{\minimaxLitePathReach}{46.4}
\newcommand{\minimaxLitePathReachHit}{26.2}

\newcommand{\minimaxLiteEM}{6.8}
\newcommand{\minimaxLitePrecision}{19.7}
\newcommand{\minimaxLiteRecall}{24.2}
\newcommand{\minimaxLiteF}{20.6}
\newcommand{\minimaxLiteNodeF}{55.3}
\newcommand{\minimaxLiteEdgeF}{29.2}

\newcommand{\minimaxLiteSqlExec}{86.9}

\newcommand{\gptFiveFourLiteN}{114}
\newcommand{\gptFiveFourLiteAnySvc}{85.1}
\newcommand{\gptFiveFourLitePathReach}{58.8}

\newcommand{\gptFiveFourLiteEM}{18.4}

\newcommand{\gptFiveFourLiteF}{22.5}
\newcommand{\gptFiveFourLiteNodeF}{68.1}
\newcommand{\gptFiveFourLiteEdgeF}{45.8}

\newcommand{\gptFiveFiveLiteN}{32}
\newcommand{\gptFiveFiveLiteAnySvc}{90.6}
\newcommand{\gptFiveFiveLitePathReach}{87.5}

\newcommand{\gptFiveFiveLiteEM}{15.6}

\newcommand{\gptFiveFiveLiteF}{19.8}
\newcommand{\gptFiveFiveLiteNodeF}{70.9}
\newcommand{\gptFiveFiveLiteEdgeF}{53.3}

\newcommand{\modelCount}{11}
\newcommand{\avgAnySvc}{76.0}
\newcommand{\avgPathReach}{61.5}

\newcommand{\avgEM}{20.7}

\newcommand{\avgRecall}{33.2}
\newcommand{\avgF}{34.1}
\newcommand{\avgNodeF}{62.2}
\newcommand{\avgEdgeF}{43.4}

\providecommand{\systemCount}{3}

\providecommand{\evalCountTt}{--}
\providecommand{\evalCountOtel}{--}
\providecommand{\evalCountHr}{--}
\providecommand{\faultKindsTt}{--}
\providecommand{\faultKindsOtel}{--}
\providecommand{\faultKindsHr}{--}
\providecommand{\meanLpTt}{--}
\providecommand{\meanLpOtel}{--}
\providecommand{\meanLpHr}{--}
\providecommand{\meanLp}{--}
\providecommand{\avgCausalEdgesTt}{--}
\providecommand{\avgCausalEdgesOtel}{--}
\providecommand{\avgCausalEdgesHr}{--}

\providecommand{\glmLiteTtPathReach}{--}\providecommand{\glmLiteTtEdgeF}{--}\providecommand{\glmLiteTtNodeF}{--}
\providecommand{\glmLiteOtelPathReach}{--}\providecommand{\glmLiteOtelEdgeF}{--}\providecommand{\glmLiteOtelNodeF}{--}
\providecommand{\glmLiteHrPathReach}{--}\providecommand{\glmLiteHrEdgeF}{--}\providecommand{\glmLiteHrNodeF}{--}

\providecommand{\seedLiteTtPathReach}{--}\providecommand{\seedLiteTtEdgeF}{--}\providecommand{\seedLiteTtNodeF}{--}
\providecommand{\seedLiteOtelPathReach}{--}\providecommand{\seedLiteOtelEdgeF}{--}\providecommand{\seedLiteOtelNodeF}{--}
\providecommand{\seedLiteHrPathReach}{--}\providecommand{\seedLiteHrEdgeF}{--}\providecommand{\seedLiteHrNodeF}{--}

\title{\datasetName: From Outcome Labels to Causal Process Supervision}

\author{%
  Aoyang Fang \quad Yifan Yang \quad
  Jin'ao Shang$^*$ \quad Qisheng Lu \quad Junjielung Xu \\
  \quad Rui Wang \quad Songhan Zhang \quad Yuzhong Zhang \quad Boxi Yu$^\P$ \quad Pinjia He$^\ddagger$\thanks{$^\ddagger$Corresponding author.} \\
  The Chinese University of Hong Kong, Shenzhen \\
  {aoyangfang, yifanyang6, qishenglu, junjielongxu, 224040299,  222010549, yuzhongzhang}@link.cuhk.edu.cn \\
  $^*$jinao\_s@stu.xjtu.edu.cn \quad $^\P$boxi.yu@lero.ie \quad $^\ddagger$hepinjia@cuhk.edu.cn
}

\begin{document}

\maketitle

\begin{abstract}
Root cause analysis (RCA) poses a holistic test of LLM agentic capabilities, such as long-context understanding, multi-step reasoning, and tool use.
However, existing datasets suffer from a fundamental gap: they label only the root cause, not the propagation path connecting it to the observed symptom, which largely simplifies the task to naive pattern matching.
To support rigorous evaluation, we introduce \frameworkName{}, a step-wise labeling protocol that leverages known interventions from fault injection to reconstruct causal propagation paths.
The mechanism is \emph{forward verification}: reasoning from cause to effect rather than inferring backward from symptoms.
Applying \frameworkName{} yields \datasetName{} (\totalValidLite{} instances), the first cross-system RCA benchmark with step-wise causal annotations for LLM agents.
Across \modelCount{} frontier LLMs, recovering the exact root-cause set succeeds in only \avgEM{}\% of cases on average.
To locate where this difficulty lies, we relax the criterion and find what we call the \emph{ungrounded diagnosis}: agents identify at least one correct root-cause service in \avgAnySvc{}\% of cases but ground that service in a verified causal propagation path to the observed symptom in only \avgPathReach{}\%.
Outcome-only evaluation hides this failure mode; step-wise causal ground truth is the missing piece for trustworthy LLM-based RCA agents.

\end{abstract}

\section{Introduction}
\label{sec:introduction}

\emph{Root cause analysis} (RCA), pinpointing which component caused a failure in a large software system, has emerged as a demanding testbed for LLM agents~\cite{xu2025openrca, chen2025aiopslab, chen2025stratus, zhang2025thinkfl}.
The originating fault is rarely where symptoms first appear: modern cloud applications are split into dozens of small components communicating over the network~\cite{zhou2018fault}, so a local fault can cascade through callers and reach distant parts of the system before any human notices.
From this cascade, the agent receives heterogeneous telemetry: request traces showing how each request propagated, time-series metrics, and free-form logs.
It must reason backward from this telemetry to the originating fault.


\begin{figure*}[t]
\centering
\includegraphics[width=\linewidth]{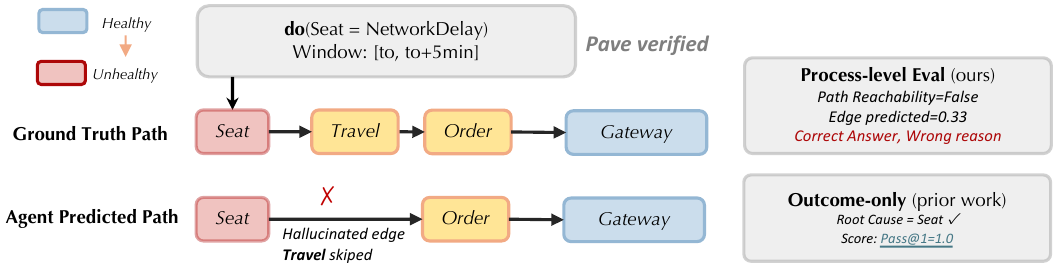}

\caption{An \emph{ungrounded diagnosis} on the running \textsf{NetworkDelay} failure in \textsf{Seat} service.
From the known intervention (top), \frameworkName{} reconstructs the verified causal path \textsf{Seat}$\to$\textsf{Travel}$\to$\textsf{Order}$\to$\textsf{Gateway} (middle).
The agent names the correct root cause but produces a graph that skips \textsf{Travel} (bottom): outcome-only evaluation scores this as a success, whereas process-level evaluation surfaces the missing edge.}
\label{fig:overview}
\end{figure*}

To evaluate such reasoning, all widely used RCA benchmarks build labeled data through \emph{fault injection}: the procedure deliberately breaks a known component (e.g., throttling its CPU, delaying its network, crashing its process), records the resulting telemetry, and treats the broken component as the ground-truth label~\cite{pham2025rcaeval, fang2025rethinking, chen2025aiopslab}.
Evaluation then scores only the outcome: whether the agent identified the correct component (Figure~\ref{fig:overview}).
The derivation that produced this answer is not examined.

Scoring the derivation rather than just the outcome reveals what we call the \emph{ungrounded diagnosis}: the agent identifies a correct root-cause service without a verified causal propagation path linking it to the observed failure.
Detecting it at scale requires step-wise causal ground truth, but building it runs into the agent's problem: from telemetry alone, an annotator would still have to reason backward from cascading effects to the originating fault, since many faults can produce similar telemetry.
But fault-injection records hand the annotator one thing the agent never sees: the intervention that started the cascade~\cite{pearl2009causality}.
This information asymmetry~\cite{liu2026self} turns the annotator's task from inference into \emph{forward verification}: with the cause given, check which downstream effects actually followed.

We turn this insight into \frameworkName{} (Path Annotation via Verified Effects), an annotation protocol that reconstructs verified causal propagation paths from fault-injection recordings.
No single signal is sufficient: a structurally plausible chain may not actually have carried the failure, a statistical co-occurrence may be coincidental, and aligned timing alone does not imply a causal link.
\frameworkName{} therefore admits a candidate path only when it satisfies all three conditions jointly: conformance to known propagation mechanisms (structural), statistically significant deviation from a pre-injection baseline (statistical), and upstream-to-downstream timing alignment (temporal).
Admitted paths thus carry consistent structural, statistical, and temporal evidence (Section~\ref{sec:phase2_verification}).

Applying \frameworkName{} across \systemCount{} architecturally distinct microservice systems (TrainTicket, the OpenTelemetry Demo, and DeathStarBench Hotel Reservation) and \faultTypeCount{} fault types yields \datasetName{}, the first cross-system RCA benchmark with step-wise causal annotations: \totalValidLite{} evaluable instances carrying \avgCausalEdges{} verified causal edges on average.

On \datasetName{}, the gap between identifying a correct service and grounding it in a verifiable path is substantial across frontier LLMs.
Averaged across \modelCount{} such models, agents recover the exact root-cause set in only \avgEM{}\% of cases.
To locate where this difficulty lies, we relax the criterion: identifying at least one correct root-cause service rises to \avgAnySvc{}\%, but grounding that service in a verified causal propagation path to the observed symptom is satisfied in only \avgPathReach{}\%, leaving roughly one in five correct identifications ungrounded.
Outcome-only evaluation hides this failure mode; step-wise causal ground truth is the missing piece for trustworthy LLM-based RCA agents.

\section{\frameworkName{}: \textbf{P}ath \textbf{A}nnotation via \textbf{V}erified \textbf{E}ffects}
\label{sec:methodology}

\begin{figure}[t]
\centering
\includegraphics[width=0.43\linewidth]{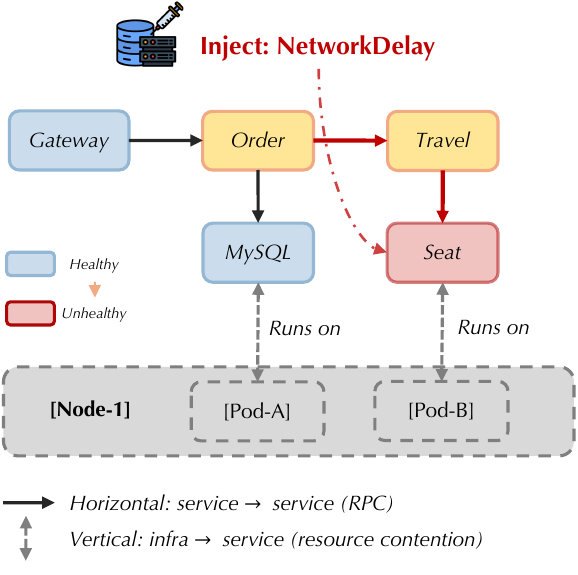}\hfill
\includegraphics[width=0.4\linewidth]{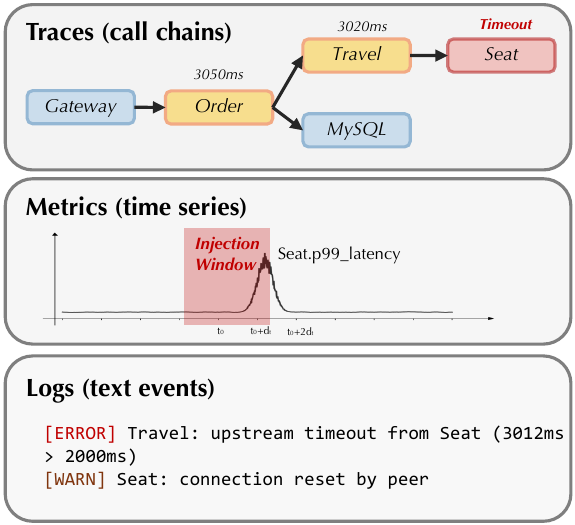}
\caption{The RCA setting.
\textbf{(a)}~A microservice system forms a two-layer dependency graph.
A fault injected at \textsf{Seat} cascades along the RPC chain
($\textsf{Seat}\to\textsf{Travel}\to\textsf{Order}$) and may also propagate vertically through shared infrastructure (e.g., co-located pods).
\textbf{(b)}~The agent observes only telemetry (traces, metrics, logs) and must reason backward to the originating fault; the intervention itself is hidden.}
\label{fig:rca_setting}
\end{figure}

A microservice system runs as a directed dependency graph over services, pods, and hosts (Figure~\ref{fig:rca_setting}a), and a fault at one node propagates along two distinct channels: horizontally along RPC call chains, and vertically across shared infrastructure.
The diagnosing agent sees only the telemetry these channels leave behind (traces, metrics, and logs; Figure~\ref{fig:rca_setting}b) and must reason backward to the originating fault.

%

\begin{figure*}[t]
\centering
\includegraphics[width=\textwidth]{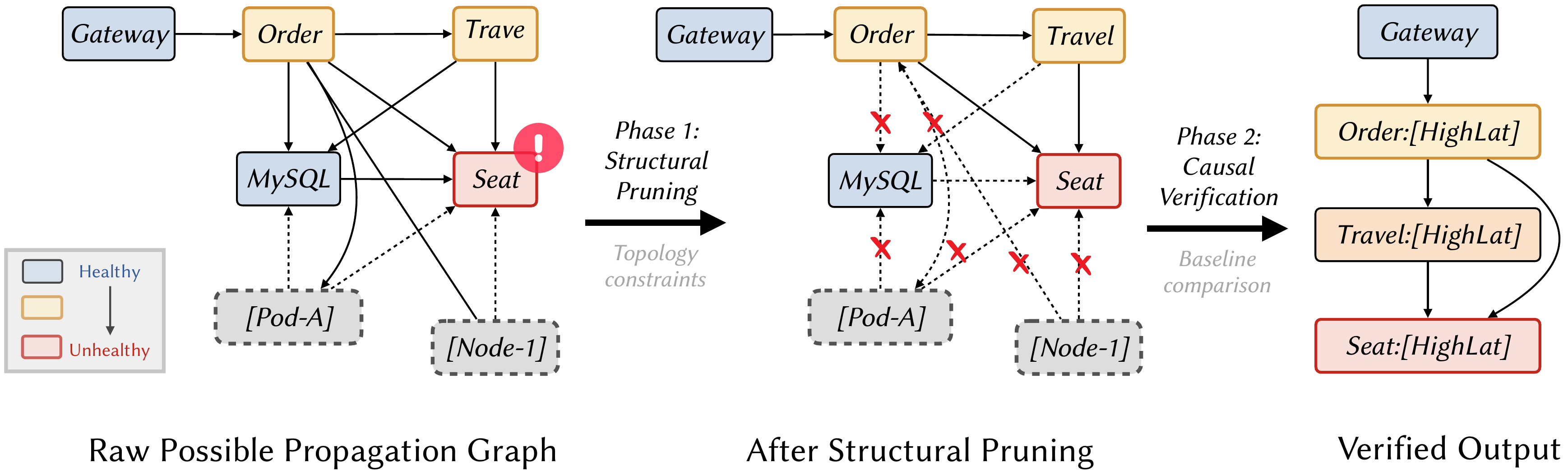}
\caption{\textbf{Coarse-to-Fine verification refines a cluttered observation graph into one verified causal chain.}
Running TrainTicket example: the true root cause is \textit{Seat}, while \textit{Route} is a benign service that appears anomalous due to background noise.
\emph{Input.} Every observed anomaly and every potential edge are kept, so the root cause cannot be picked out.
\emph{Phase~1 (Structural Pruning).} Edges marked $\times$ are removed because the topology admits no such dependency (e.g.,~\textit{Route}$\rightarrow$\textit{Seat}); surviving solid edges are structurally possible, but \textit{Route} itself is still flagged.
\emph{Phase~2 (Causal Verification).} Each candidate is scored against a pre-injection baseline; \textit{Route}'s deviation matches baseline noise and is dropped, leaving the chain \textit{Seat}$\rightarrow$\textit{Travel}$\rightarrow$\textit{Order}.
Phase~1 asks ``\emph{could this edge happen?}'' (topology); Phase~2 asks ``\emph{did this anomaly carry the signal?}'' (baseline contrast); injections with no surviving chain are discarded, and the released artifact is the union of such chains.
}
\label{fig:pipeline}
\end{figure*}

We now describe the \frameworkName{} pipeline (Figure~\ref{fig:pipeline}).
Given fault injection records and the system dependency graph, it reconstructs verified causal propagation paths in two phases that together enforce three conditions on every admitted path.
\textit{Structural Pruning} (Section~\ref{sec:phase1_pruning}) enumerates candidate paths consistent with the topology and a propagation rule set, enforcing \emph{mechanism conformance}.
\textit{Causal Verification} (Section~\ref{sec:phase2_verification}) then checks each candidate against pre-injection telemetry, enforcing \emph{statistical deviation} and \emph{timing alignment} jointly.

For each injection, \frameworkName{} consumes three inputs (the dependency graph $\mathcal{G}$, raw telemetry $\mathbf{O}^{(t)}$ over the injection window, and the recorded intervention $do(v_{root})$) and emits one output: a verified path set $\Pi^*$.
Phase~1 builds an intermediate candidate set $\Pi_{cand}$ from topology and the rule set; Phase~2 yields $\Pi^* \subseteq \Pi_{cand}$ by verifying survivors on telemetry.
Each injection in \datasetName{} carries $\Pi^*$ as its step-wise annotation.

\subsection{Setup: Forward Verification from a Known Intervention}
\label{sec:preliminaries}

We model a microservice system as a directed dependency graph $\mathcal{G} = (\mathcal{V}, \mathcal{E})$ over services, pods, and hosts (Figure~\ref{fig:rca_setting}a), with edges encoding invocation or resource dependencies (channels and modalities in Appendix~\ref{sec:appendix_channels_modalities}).
Fault propagation is a structural causal process over $\mathcal{G}$~\cite{pearl2009causality}: anomalies propagate along edges with finite delay (full equation in Eq.~\eqref{eq:scm}, Appendix~\ref{sec:appendix_scm}).
An RCA agent observes only telemetry $\mathbf{O}_{obs}$ and must jointly infer the dependency structure, the root cause, and the propagation path: an inherently \emph{backward} abductive problem.
\frameworkName{}, by contrast, has access to the intervention $do(v_{root})$ (target component, fault type, injection window $[t_0, t_0{+}\Delta t]$, and parameters), giving it a strictly richer information set:
\begin{equation}
    \underbrace{\{\mathbf{O}_{obs}\}}_{\text{Agent}}
    \;\subset\;
    \underbrace{\{\mathbf{O}_{obs},\, do(v_{root})\}}_{\text{\frameworkName{}}}
    \label{eq:info_asymmetry}
\end{equation}
Knowing the intervention turns an ill-posed inverse problem into a well-posed \emph{forward} verification task: confirming that observed anomalies are consistent with the known intervention propagating through $\mathcal{G}$.
The verified causal subgraph $\mathcal{G}^*$ (formed by the edges of all paths in $\Pi^*$) is the process-level ground truth (GT) against which agent reasoning is evaluated.

\subsection{Phase 1: Structural Pruning}
\label{sec:phase1_pruning}

Phase 1 produces a candidate path set $\Pi_{cand}$ by enforcing topological constraints from the system dependency graph.
Raw telemetry $\mathbf{O}^{(t)}$ is first projected onto a discrete state alphabet $\Sigma$ via a node-type-aware mapping $\Psi$ (e.g., \texttt{HighLatency} for service nodes, \texttt{CPU\_Throttled} for pods; formal definition in Appendix~\ref{sec:appendix_state_abstraction}).
A propagation rule set $\mathcal{R}$ admits only transitions between (type, state) pairs $(\theta_i{:}s_i) \xrightarrow{\text{channel}} (\theta_j{:}s_j)$ backed by a known mechanism (resource contention, RPC timeout cascade, etc.).
A constrained DFS over a time-expanded heterogeneous graph, whose nodes are (component, state, time) triples $(v, s, t)$, unrolls circular dependencies into acyclic chains, yielding $\Pi_{cand}$.
The state alphabet, full rule table, and time-expansion details are in Appendix~\ref{sec:appendix_pipeline_details}.

\subsection{Phase 2: Causal Verification}
\label{sec:phase2_verification}

Phase 2 produces the verified path set $\Pi^*$ by separating genuine propagation from coincidental disturbances (e.g., periodic cron jobs) that co-occur with the injection window.
For each injection, we collect normal-period telemetry on a fresh, warmed-up instance as a \textit{Reference Baseline} ($\mathcal{D}_{base}$) immediately before injection, approximating $P(\cdot \mid do(v_{root}{=}\emptyset))$.
For each node $v_i$ along a candidate path $\pi \in \Pi_{cand}$, we apply distributional tests tailored to each metric type (Z-score for Gaussian metrics, percentile-based tests for heavy-tailed metrics; Appendix~\ref{sec:appendix_thresholds}).
A candidate path $\pi$ is \emph{verified} only if \emph{all} of (i)~every node along $\pi$ exhibits a deviation beyond its baseline screen, (ii)~every consecutive edge conforms to $\mathcal{R}$ (inherited from Phase~1), and (iii)~each downstream anomaly onsets no earlier than its upstream cause hold simultaneously.

Phase 2 admits an injection into the released artifact only when its cascade clears the joint screen end-to-end: the fault must produce an observable downstream signature (a deviation crossing the configured SLO surface), at least one rule-admitted, temporally consistent path must connect the intervention to that signature, and the pre-injection baseline must be free of drift.
Cases failing any of these conditions are dropped during pipeline-level analysis.

\section{Experiments}
\label{sec:evaluation}

Process-level ground truth reveals a reasoning gap in LLM-based RCA agents that outcome-only evaluation hides: across \modelCount{} frontier LLMs, correct root-cause identifications are routinely not grounded in any verified causal propagation path.

\subsection{Experimental Setup}
\label{sec:exp_setup}

This subsection specifies the three microservice systems we run, the fault-injection campaigns that produce the raw pool, the curation that yields \datasetName{}, and the agent design under evaluation.

\textbf{Systems and Faults.}
We evaluate on \systemCount{} architecturally distinct microservice systems: \emph{TrainTicket}~\cite{zhou2018fault}, \emph{Hotel Reservation} from DeathStarBench~\cite{DeathStarBench}, and the \emph{OpenTelemetry Demo} (OTel Demo) application~\cite{OpenTelemetryDemo}, spanning different language stacks and synchronous/asynchronous communication patterns; deployment specifications are in Appendix~\ref{sec:appendix_systems}.
For each system we run fault-injection workloads covering \faultTypeCount{} chaos primary kinds across application, resource, network, and pod-level faults (per-system taxonomies in Appendix~\ref{sec:appendix_fault_types}); for TrainTicket we build on the released dataset of Fang et al.~\cite{fang2025rethinking} and augment it with new campaigns on the other two systems.
The resulting raw pool comprises \totalRawInjections{} injected runs.
After filtering silent injections, we apply the \frameworkName{} pipeline (Section~\ref{sec:methodology}) to the survivors and curate \datasetName{} (\totalValidLite{} instances) by stratified selection on (system, chaos family, root service) with per-stratum caps and additional structural screens (full rules in Appendix~\ref{sec:appendix_datasheet}).
Table~\ref{tab:dataset_summary} reports per-system breakdowns; the Datasheet is in Appendix~\ref{sec:appendix_datasheet}.

\begin{table}[t]
\centering
\caption{\textbf{\datasetName{} by system.}
The three systems differ in size (9--44 services), language stack, and typical propagation depth, so the \totalValidLite{} instances cover distinct architectures rather than re-sampling one.
Cluster configuration in Appendix~\ref{sec:appendix_cluster}; raw pool sizes and silent-injection rates in Appendix~\ref{sec:appendix_datasheet}.}
\label{tab:dataset_summary}
\resizebox{0.8\columnwidth}{!}{%
\begin{tabular}{l rrr r}
\toprule
\textbf{Property} & \textbf{TrainTicket} & \textbf{OTel Demo} & \textbf{Hotel Res.} & \textbf{Total} \\
\midrule
Services in system              & 44               & 15                  & 9               &     \\
Implementation                  & Java/REST        & polyglot/gRPC+Kafka & Go/gRPC         &     \\
\midrule
Evaluable instances             & \evalCountTt     & \evalCountOtel      & \evalCountHr    & \totalValidLite \\
Distinct chaos primary kinds    & \faultKindsTt    & \faultKindsOtel     & \faultKindsHr   & \faultTypeCount \\
Mean longest propagation path   & \meanLpTt        & \meanLpOtel         & \meanLpHr       & \meanLp \\
Avg.\ causal edges per instance & \avgCausalEdgesTt & \avgCausalEdgesOtel & \avgCausalEdgesHr & \avgCausalEdges \\
\bottomrule
\end{tabular}%
}
\end{table}

\textbf{Ground Truth.}
Causal propagation paths are extracted automatically using the pipeline described in Section~\ref{sec:methodology}; their reliability is bounded by the two-annotator manual audit reported in Section~\ref{sec:exp_gt_validation}.

\textbf{Agent Design.}
All evaluated agents follow a tool-augmented ReAct-style~\cite{yao2022react} architecture, adopted from the Deep Research agent~\cite{lin2025deepresearch}.
To isolate LLM reasoning capability from tooling effects, all models share an identical tool suite over the released telemetry and identical prompt templates; no dependency graph is provided.
Each diagnosis is returned as structured output: root-cause claims and propagation edges, each paired to a re-executable SQL query as evidence (full schema, tool definitions, and prompts in Appendix~\ref{sec:appendix_baselines} and~\ref{sec:appendix_causalgraph}).

\subsection{Benchmark Utility: Process-Level Evaluation of RCA Agents}
\label{sec:exp_benchmark}

This subsection grades \modelCount{} frontier LLMs against \datasetName{} along outcome and process layers and reads three patterns out of the pooled results: outcome scores understate the failure (Finding~1), process-level Edge F1 is universally lower than Node F1 (Finding~2), and no single backbone dominates (Finding~3).

\textbf{Experimental Design.}
We evaluate the RCA agent architecture above with \modelCount{} frontier LLMs as backbones spanning closed-source and open-source families across Anthropic, Google, ByteDance, Xiaomi, Zhipu, Moonshot, DeepSeek, Alibaba, Tencent, and MiniMax: Claude Opus 4.7, Gemini 3.1 Pro, Claude Sonnet 4.6, MiMo 2.5 Pro, GLM 5.1, Kimi K2.6, DeepSeek V4 Pro, Qwen3.6-Max, Hy 3.0 Preview, Seed 2.0 Pro, and MiniMax M2.7.
Each model receives multi-modal telemetry (traces, metrics, logs) and must diagnose the root cause while explaining the failure propagation as a causal graph.

\textbf{Evaluation Metrics.}
We grade each diagnosis along two complementary layers (formal definitions in Appendix~\ref{sec:appendix_metrics}).
The \textit{outcome layer} scores the agent's predicted root-cause set against ground truth: \textit{Exact Match} (EM) requires the predicted set to match ground truth exactly; \textit{F1}, \textit{Precision}, and \textit{Recall} are mean fractional scores over the (service, fault\_kind) pair set; \textit{AnySvc} is the case-level rate of naming at least one correct root-cause service ignoring fault kind, the most lenient outcome score and a service-only relaxation of Recall.
The \textit{process layer} scores the agent's reasoning: \textit{Path Reachability} (PR) is satisfied when the agent both names at least one correct root-cause service \emph{and} draws a valid path through its predicted causal graph from that service to a ground-truth alarm node, a component observed to behave abnormally during the injection window; \textit{Node F1} and \textit{Edge F1} score the predicted causal graph $\hat{\mathcal{G}}$ against $\mathcal{G}^*$ under exact matching; \textit{SQL Exec} is the share of agent-emitted evidence SQL queries that execute cleanly.
PR is the most lenient process metric: it requires only one valid path from a single correct anchor, not the full graph.
We reserve the term \emph{ungrounded diagnosis} for cases that satisfy AnySvc but not PR: a correct root-cause identification not grounded in a verified causal propagation path to the observed symptom.

\textbf{Results.}
\begin{table}[!t]
\centering
\caption{Outcome- and process-level evaluation on \datasetName{} pooled across \systemCount{} systems ($n{=}\totalValidLite{}$, percentages).
\textbf{Bold} marks the per-column maximum.
Metric definitions in Section~\ref{sec:exp_benchmark}; per-system breakdown in Appendix Table~\ref{tab:llm_evaluation_per_system}.}
\label{tab:llm_evaluation}
\resizebox{\textwidth}{!}{%
\begin{tabular}{l ccccc cccc}
\toprule
 & \multicolumn{5}{c}{\textit{Outcome}} & \multicolumn{4}{c}{\textit{Process}} \\
\cmidrule(lr){2-6} \cmidrule(lr){7-10}
Model & EM$\uparrow$ & F1$\uparrow$ & Precision$\uparrow$ & Recall$\uparrow$ & AnySvc$\uparrow$ & PR$\uparrow$ & Node F1$\uparrow$ & Edge F1$\uparrow$ & SQL Exec$\uparrow$ \\
\midrule
Claude Opus 4.7              & \opusLiteEM             & \opusLiteF             & \opusLitePrecision     & \opusLiteRecall             & \textbf{\opusLiteAnySvc}    & \opusLitePathReach          & \opusLiteNodeF             & \opusLiteEdgeF             & \opusLiteSqlExec             \\
Gemini 3.1 Pro               & \textbf{\geminiLiteEM}  & \textbf{\geminiLiteF}  & \textbf{\geminiLitePrecision} & \textbf{\geminiLiteRecall} & \geminiLiteAnySvc           & \geminiLitePathReach        & \textbf{\geminiLiteNodeF}  & \geminiLiteEdgeF           & \textbf{\geminiLiteSqlExec}  \\
Claude Sonnet 4.6            & \sonnetLiteEM           & \sonnetLiteF           & \sonnetLitePrecision   & \sonnetLiteRecall           & \sonnetLiteAnySvc           & \sonnetLitePathReach        & \sonnetLiteNodeF           & \sonnetLiteEdgeF           & \sonnetLiteSqlExec           \\
MiMo 2.5 Pro                 & \mimoLiteEM             & \mimoLiteF             & \mimoLitePrecision     & \mimoLiteRecall             & \mimoLiteAnySvc             & \mimoLitePathReach          & \mimoLiteNodeF             & \mimoLiteEdgeF             & \mimoLiteSqlExec             \\
GLM 5.1                      & \glmLiteEM              & \glmLiteF              & \glmLitePrecision      & \glmLiteRecall              & \glmLiteAnySvc              & \glmLitePathReach           & \glmLiteNodeF              & \textbf{\glmLiteEdgeF}     & \glmLiteSqlExec              \\
Kimi K2.6                    & \kimiLiteEM             & \kimiLiteF             & \kimiLitePrecision     & \kimiLiteRecall             & \kimiLiteAnySvc             & \kimiLitePathReach          & \kimiLiteNodeF             & \kimiLiteEdgeF             & \kimiLiteSqlExec             \\
DeepSeek V4 Pro              & \dsvfourLiteEM          & \dsvfourLiteF          & \dsvfourLitePrecision  & \dsvfourLiteRecall          & \dsvfourLiteAnySvc          & \dsvfourLitePathReach       & \dsvfourLiteNodeF          & \dsvfourLiteEdgeF          & \dsvfourLiteSqlExec          \\
Qwen3.6-Max                  & \qwenLiteEM             & \qwenLiteF             & \qwenLitePrecision     & \qwenLiteRecall             & \qwenLiteAnySvc             & \textbf{\qwenLitePathReach} & \qwenLiteNodeF             & \qwenLiteEdgeF             & \qwenLiteSqlExec             \\
Hy 3.0 Preview               & \hyLiteEM               & \hyLiteF               & \hyLitePrecision       & \hyLiteRecall               & \hyLiteAnySvc               & \hyLitePathReach            & \hyLiteNodeF               & \hyLiteEdgeF               & \hyLiteSqlExec               \\
Seed 2.0 Pro                 & \seedLiteEM             & \seedLiteF             & \seedLitePrecision     & \seedLiteRecall             & \seedLiteAnySvc             & \seedLitePathReach          & \seedLiteNodeF             & \seedLiteEdgeF             & \seedLiteSqlExec             \\
MiniMax M2.7                 & \minimaxLiteEM          & \minimaxLiteF          & \minimaxLitePrecision  & \minimaxLiteRecall          & \minimaxLiteAnySvc          & \minimaxLitePathReach       & \minimaxLiteNodeF          & \minimaxLiteEdgeF          & \minimaxLiteSqlExec          \\
\bottomrule
\end{tabular}%
}
\end{table}

Three patterns govern the pooled results in Table~\ref{tab:llm_evaluation}.
First, outcome-level scores are already low (mean F1 reaches only $\avgF{}\%$ and EM only $\avgEM{}\%$ across the \modelCount{} models), yet outcome alone obscures the underlying failure: agents name a correct service in $\avgAnySvc{}\%$ of cases (AnySvc), and the gap between this lenient outcome score and the most lenient process score ($\avgPathReach{}\%$ PR) reveals what we call the \emph{ungrounded diagnosis} (Finding~1).
Second, within the process layer mean Edge F1 ($\avgEdgeF{}\%$) trails mean Node F1 ($\avgNodeF{}\%$) by \fpeval{\avgNodeF-\avgEdgeF}\,pp for every model, indicating that agents identify participating services more reliably than the directed dependencies among them (Finding~2).
Third, the per-column extrema are distributed across four different models with distinct recall and precision profiles, so no single backbone dominates and the appropriate ranking depends on the downstream verification cost (Finding~3).
The per-system spread is wider than the cross-model spread (Hotel Reservation approaches saturation on the outcome layer for the top tier, whereas TrainTicket leaves a 30+\,pp gap on the process-level metrics); we therefore keep the main table pooled to support a single comparable model ranking and report the per-system breakdown in Appendix Table~\ref{tab:llm_evaluation_per_system}.

\textbf{(1) Outcome scores understate the failure: the lenient process check exposes ungrounded diagnoses.}
The outcome layer pools two distinct failure modes into one number.
On the lenient end, AnySvc reaches $\avgAnySvc{}\%$ on average, showing that agents routinely name a correct root-cause service.
The strict outcome metrics, however, collapse: pair-level F1 falls to $\avgF{}\%$ and EM to $\avgEM{}\%$, with the drop from AnySvc ($\avgAnySvc{}\%$) to Recall ($\avgRecall{}\%$) dominated by mislabelled fault kinds on services the agent did identify.
The decisive evidence that this gap is a \emph{reasoning} failure rather than a labelling artefact comes from the process layer: even PR, which requires only one valid path from a single correct anchor, is satisfied in only $\avgPathReach{}\%$ of cases, so for \fpeval{\avgAnySvc-\avgPathReach}\,pp of cases the agent names a correct service \emph{without} drawing any verifiable path to the observed symptom.
The per-model conditional rate $\text{PR}/\text{AnySvc}$ separates models that outcome-only grading rates as comparable: Qwen3.6-Max attains $\qwenLitePathReach{}/\qwenLiteAnySvc{} \approx \fpeval{round(\qwenLitePathReach/\qwenLiteAnySvc*100,0)}\%$, whereas Sonnet 4.6 attains only $\sonnetLitePathReach{}/\sonnetLiteAnySvc{} \approx \fpeval{round(\sonnetLitePathReach/\sonnetLiteAnySvc*100,0)}\%$ despite comparable F1, indicating that Sonnet emits correct service labels that are not integrated into a connected propagation path.

\textbf{(2) Edge F1 is universally lower than Node F1, and the gap is the size of the entire process-level signal.}
The ordering holds for every model in the table, and the gap is substantial in absolute terms ($\avgNodeF{}\%$ vs.\ $\avgEdgeF{}\%$, \fpeval{\avgNodeF-\avgEdgeF}\,pp on average).
Gemini exhibits the widest split, reaching $\geminiLiteNodeF{}\%$ Node F1 but only $\geminiLiteEdgeF{}\%$ Edge F1.
Identifying the participating services approximates a classification task that pre-training prepares the model for; ordering them into a directed propagation path requires reasoning about what each abnormal signal implies for the underlying inter-service dependencies, a derivation step that outcome-only metrics do not score.
This separation provides direct evidence that the agent's \emph{output} (a service list) is closer to ground truth than its \emph{derivation} (a directed graph), and that the discrepancy is large enough to affect model selection in practice.


\subsection{Failure Mode Characterization}
\label{sec:exp_failure_modes}

To probe what drives the shortfall between AnySvc and PR, we read agent trajectories across the \modelCount{} models on a stratified sample of cases tiered by joint correctness (the fraction of models clearing $F_1{>}0$ on each case).
Three patterns recur across the sample, each describing how an agent commits to an incorrect or unsupported answer rather than continuing to verify, and together they account for the dominant share of ungrounded diagnoses.

\textbf{(1) Premature commitment under under-exploration.}
After identifying one plausible root cause, the agent stops investigating and emits the answer, leaving co-occurring faults or alternative attributions unexamined.
This shape is most visible on hybrid injections that pair one prominent fault with one subtle one: every evaluated model converges on the prominent half (often emitting nearly identical wording across heterogeneous backbones) and uniformly misses the subtle half, depressing Recall and EM whenever the ground truth covers more than one service.

\textbf{(2) Presence bias on observable activity.}
Agents weight evidence by what they see and treat absence of observed activity as evidence of normal operation, rather than recognizing that the absence itself can be the fault signature.
A killed-pod injection makes this failure mode crisp: during the incident window the affected service stops producing spans entirely, and many agents read the resulting silence as ``this service has no errors, hence no fault here'', then attribute the incident to the upstream callers that surface errors precisely because their callee disappeared.
The dual of Pattern~(3): attentional bias toward strong positive signals and against silent ones jointly explain the bulk of the localization shortfall.

\textbf{(3) Salience capture on the loudest signal.}
Agents converge on the service whose error or latency aggregates carry the strongest abnormal-versus-normal delta, even when those signals are pure downstream amplification rather than the originating fault; this pattern accounts for the bulk of the localization failures responsible for AnySvc falling short of full coverage by roughly $24$pp.
The capture operates on aggregate metrics (a high-traffic hub on the load-generator path absorbs attention while the lower-traffic ground-truth service goes unnamed) and equally on text: a SEVERE log line claiming a timeout is taken at face value by most models even when the corresponding trace span returns in tens of milliseconds and contradicts the log claim.

The case study below illustrates Pattern~(1) on the path-drawing dimension specifically: an agent commits to the correct service and stops verifying, leaving the propagation chain to skip the intermediates that the verified ground truth actually traverses.

\textbf{Case Study: An Ungrounded Diagnosis.}
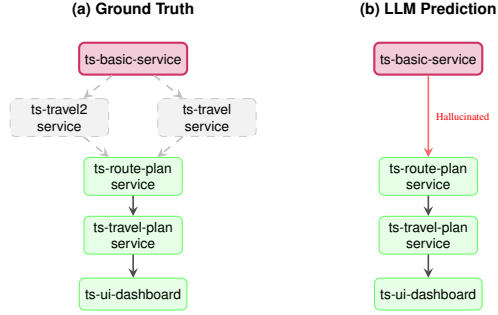
\begin{figure}[t]
\centering
\begin{tikzpicture}[
    scale=0.65, transform shape,
    service/.style={rectangle, draw=black!60, fill=white, minimum width=2cm, minimum height=0.65cm, font=\footnotesize\sffamily, rounded corners=2pt, align=center},
    root/.style={service, fill=purple!20, draw=purple!80, line width=0.8pt},
    correct/.style={service, fill=green!10, draw=green!60},
    missed/.style={service, fill=gray!10, draw=gray!40, dashed},
    edge_correct/.style={->, >=stealth, semithick, black!70},
    edge_missed/.style={->, >=stealth, semithick, gray!50, dashed},
    edge_wrong/.style={->, >=stealth, semithick, red!60},
    label_text/.style={font=\small\bfseries\sffamily, align=center}
]

\node[label_text] at (0, 1) {(a) Ground Truth};

\node[root] (gt_basic) at (0, 0) {ts-basic-service};

\node[missed] (gt_travel2) at (-1.5, -1.2) {ts-travel2\\service};
\node[missed] (gt_travel) at (1.5, -1.2) {ts-travel\\service};

\node[correct] (gt_route) at (0, -2.4) {ts-route-plan\\service};
\node[correct] (gt_plan) at (0, -3.6) {ts-travel-plan\\service};
\node[correct] (gt_ui) at (0, -4.8) {ts-ui-dashboard};

\draw[edge_missed] (gt_basic) -- (gt_travel2);
\draw[edge_missed] (gt_basic) -- (gt_travel);
\draw[edge_missed] (gt_travel2) -- (gt_route);
\draw[edge_missed] (gt_travel) -- (gt_route);
\draw[edge_correct] (gt_route) -- (gt_plan);
\draw[edge_correct] (gt_plan) -- (gt_ui);

\node[label_text] at (6, 1) {(b) LLM Prediction};

\node[root] (llm_basic) at (6, 0) {ts-basic-service};
\node[correct] (llm_route) at (6, -2.4) {ts-route-plan\\service};
\node[correct] (llm_plan) at (6, -3.6) {ts-travel-plan\\service};
\node[correct] (llm_ui) at (6, -4.8) {ts-ui-dashboard};

\draw[edge_wrong] (llm_basic) -- node[midway, right, font=\tiny, text=red, xshift=1pt] {Hallucinated} (llm_route);
\draw[edge_correct] (llm_route) -- (llm_plan);
\draw[edge_correct] (llm_plan) -- (llm_ui);

\end{tikzpicture}
\caption{Spurious causal reasoning case. (a) Ground truth shows fault propagation through intermediate services (\texttt{travel}, \texttt{travel2}). (b) Claude correctly identifies root cause and symptoms but hallucinates a direct edge, missing the branching structure (Edge F1=44.4\%).}
\label{fig:case_lucky_guess}
\end{figure}

Figure~\ref{fig:case_lucky_guess} shows an injection where Claude correctly names the root cause (\textsf{ts-basic-service}) but justifies it with a graph that skips the services actually on the 4-hop ground-truth propagation path: a paradigmatic ungrounded diagnosis.
The predicted graph hallucinates a direct edge from \textsf{ts-basic-service} to \textsf{ts-route-plan-service}, skipping the intermediate \textsf{ts-travel2-service} and \textsf{ts-travel-service} (Node Precision/Recall 100\%/66.7\%, Edge Precision/Recall 66.7\%/33.3\%, Edge F1 44.4\%).
Outcome-only evaluation scores this as a hit on AnySvc; process-level evaluation surfaces the missing edges and exposes the reasoning failure an on-call engineer would still hit in practice.

\subsection{Annotation Reliability}
\label{sec:exp_gt_validation}

A two-annotator manual audit on $N{=}100$ instances drawn uniformly at random from \datasetName{} reports $94\%$ case-level agreement ($94/100$) on the validity of $\mathcal{G}^*$, with every disagreement attributable to under-coverage of weakly supported paths rather than to a spurious included edge.
Two of the paper's authors, with prior experience in distributed-system observability and the \frameworkName{} schema, independently rated each case as \textsc{Agree} when every edge in $\mathcal{G}^*$ is plausibly causal from the telemetry and no obvious propagation edge is missing, and \textsc{Disagree} otherwise; annotators were blind to one another's verdicts.
All six disagreements were one-directional: at least one annotator considered an additional propagation path plausible that the verification gates had rejected as weakly supported, and no case was flagged for an implausible included edge.
Joint adjudication attributed every under-coverage instance to the conjunctive design's deliberate strictness on borderline evidence; for a process-level reference, omitting weakly supported edges is preferable to admitting spurious ones, so we accept this precision-favoured trade-off.
The full protocol, verification procedure, and per-disagreement notes are in Appendix~\ref{sec:appendix_audit}.

\section{Related Work}
\label{sec:related_work}

\subsection{From Classification-Based Ranking to Generative Reasoning}
The formulation of Root Cause Analysis (RCA) has undergone a fundamental shift.
Traditionally, RCA was cast as a \textit{multi-class classification problem}, where each candidate component is scored and ranked by its likelihood of being the root cause, using heuristic algorithms (e.g., Random Walk~\cite{yu2021microrank}) or supervised graph neural networks~\cite{zhang2025dynacausal}.
While effective for locating faults, these ``black-box'' approaches lack interpretability and fail to capture the sequential propagation mechanism.
With the advent of Large Language Models (LLMs), recent works have pivoted towards a \textit{generative reasoning paradigm}~\cite{xu2025openrca, chen2025aiopslab, chen2025stratus, zhang2025thinkfl}, treating RCA as a causal narrative generation task.
However, a critical misalignment persists: while the model architecture has evolved to handle complex reasoning, the underlying data supervision remains stagnant.
Existing benchmarks~\cite{pham2025rcaeval,fang2025rethinking,chen2025aiopslab} still rely on outcome-based labels that record only \textit{which} component is the root cause, without annotating \textit{how} the fault propagates, inherited from the classification era.
This forces reasoning agents to optimize for the final answer directly, bypassing the logical derivation steps and encouraging the memorization of spurious correlations rather than learning robust causal dynamics.

\subsection{Causal Structure Learning under Intervention}
Our work intersects with causal discovery, particularly in dynamic systems.
Classical observational methods (e.g., PC algorithm~\cite{spirtes2000causation}) struggle to recover Directed Acyclic Graphs (DAGs) in distributed systems due to unobserved confounders and high-dimensional noise~\cite{ikram2022root, han2025root}.
Pearl's hierarchy of causation posits that interventional data ($do(x)$) is essential for distinguishing true causality from mere correlation~\cite{pearl2009causality}.
While some RCA methods attempt to infer causal graphs relying solely on observational telemetry~\cite{li2022constructing, yu2023nezha}, they lack the \textit{identifiability} guarantees that controlled interventions provide.
\frameworkName{} distinguishes itself by adopting an \textit{interventional} rather than purely observational perspective during annotation.
While prior benchmarks use only the identity of the injected component as the ground-truth label, \frameworkName{} leverages the full intervention specification ($do(v_{root})$) to actively verify propagation paths.
This effectively transforms the intractable structure learning problem into a tractable verification task, filtering out silent injections and confounding noise that plague purely observational datasets.

\subsection{Process Supervision and Verifiability}
In the broader LLM landscape, focus is shifting from outcome-based supervision (Outcome Reward Models, ORMs) to \textit{Process Supervision} (Process Reward Models, PRMs)~\cite{lightman2023let, uesato2022solving, zengversaprm}.
Research in mathematical reasoning and code generation demonstrates that providing step-by-step verification signals significantly reduces hallucinations and improves generalization~\cite{yin2025dynamic, yan2025survey}.
However, applying this to system diagnostics is hindered by the \textit{verification gap}: unlike math (where proofs are deterministic) or code (where unit tests exist), operational faults lack a natural ``compiler'' to verify reasoning steps.
Current RCA benchmarks like AIOpsLab~\cite{chen2025aiopslab} or OpenRCA~\cite{xu2025openrca} rely on human-curated scenarios or heuristically filtered logs, which are hard to scale and lack granular process labels.
\frameworkName{} addresses this verification gap for system diagnostics by synthesizing verifiable failure propagation paths rooted in system topology and causal interventions, producing the first large-scale process-level ground truth for \textit{evaluating} diagnostic reasoning in complex systems.
The resulting step-wise annotations also open a path toward training PRMs in the reliability domain, though we leave this exploration to future work.

\begin{table}[t]
\centering
\caption{Differentiation of \datasetName{} from prior RCA benchmarks. ``\checkmark'' = present; ``--'' = absent or not reported as a fixed-size benchmark; ``part.'' = partial form (e.g., issue-type tag without a propagation graph). Instance counts are reported as in each source.}
\label{tab:benchmark_comparison}
\resizebox{0.7\columnwidth}{!}{%
\begin{tabular}{lcccccc}
\toprule
Benchmark & \#Sys & \#Inst. & Outcome & \textbf{Process} & SLO surface & LLM eval \\
\midrule
OpenRCA 1.0~\cite{xu2025openrca}             & 3  & 335    & \checkmark & --                  & implicit & \checkmark \\
AIOpsLab~\cite{chen2025aiopslab}             & 2  & 48     & \checkmark & part.               & implicit & \checkmark \\
RCAEval~\cite{pham2025rcaeval}               & 3  & 735    & \checkmark & --                  & --       & --         \\
Stratus~\cite{chen2025stratus}               & 3  & --     & \checkmark & --                  & implicit & \checkmark \\
Fang et al.~\cite{fang2025rethinking}        & 1  & \totalRawInjections{} & \checkmark & --   & implicit & --         \\
\textbf{\datasetName{} (ours)}               & \systemCount{} & \totalValidLite{} & \checkmark & \textbf{\checkmark} & \textbf{explicit} & \checkmark \\
\bottomrule
\end{tabular}%
}
\end{table}

\textbf{Differentiation summary.}
Table~\ref{tab:benchmark_comparison} compares \datasetName{} to five prior RCA benchmarks on seven axes.
Two columns separate our work from prior benchmarks.
\emph{Process annotations} record the step-wise causal edges of each cascade rather than only the root-cause identity; AIOpsLab provides a partial form (issue-type tags without a propagation graph), and the others provide none.
The \emph{SLO surface as an explicit method input} promotes a threshold that prior work fixes implicitly into a configurable knob.
\section{Assumptions and Threats to Validity}
\label{sec:threats}

This section delimits \datasetName{}'s claims along three axes: the input conditions \frameworkName{} relies on, the residual threats internal to the pipeline, and the boundary of external transfer.

\textbf{Input assumptions.}
Four falsifiable conditions underwrite the protocol: a real controlled intervention $do(v_{root})$ rather than an inferred label, a service-layer dependency graph complete enough that genuine horizontal propagation survives structural pruning, a propagation signature in trace, metric, or log modalities, and a pre-injection baseline that approximates the no-intervention distribution closely enough for baseline-relative anomaly screening to discriminate.
Where any of these fails for a target setting, the reported numbers should be read as an upper bound rather than a transferable estimate.

\textbf{Internal threats.}
Two single-gate weaknesses remain inside the pipeline: the rule set $\mathcal{R}$ may silently prune propagation mechanisms outside its current vocabulary, and the per-node anomaly screen is calibrated toward low false negatives (calibration split in Appendix~\ref{sec:appendix_cal_eval}) and so admits individual nodes spuriously.
Neither propagates to end-to-end errors, because admission requires the conjunction of rule consistency, baseline-relative deviation, and upstream-to-downstream timing alignment (Eq.~\eqref{eq:conjunction_fp}), and a two-annotator audit on a uniform-random 100-case subsample reports zero false-positive edges in the audited sample (Section~\ref{sec:exp_gt_validation}).

\textbf{External transfer.}
Faithful simulation of production-scale microservice deployments is neither a goal of this benchmark nor realistically achievable inside controlled testbeds, so reported numbers apply within the testbed envelope and transfer to production must be calibrated rather than assumed.
A complementary scope note concerns information conditions: \frameworkName{}'s annotation step requires full telemetry, since forward verification relies on it, but the evaluation step does not.
The same step-wise ground truth, established once under full data, can be reused to evaluate agents under deliberately restricted telemetry (subsampled traces, missing modalities, truncated observation windows), turning information availability into a controllable evaluation axis that we leave to future work.

\section{Conclusion}
\label{sec:conclusion}

We introduced \frameworkName{}, an annotation protocol that reconstructs verified causal propagation paths from fault-injection experiments by exploiting the information asymmetry between backward RCA and forward verification from a known intervention.
Its application across \systemCount{} architecturally distinct microservice systems yields \datasetName{} (\totalValidLite{} instances), the first cross-system RCA benchmark with step-wise causal annotations and process-level metrics (Path Reachability, Node F1, Edge F1) that grade the \emph{causal path} rather than only the identified root cause.
On this benchmark, process-level evaluation of \modelCount{} frontier LLMs exposes the \emph{ungrounded diagnosis}, a failure mode invisible to outcome-only evaluation in which correct root-cause identifications are not grounded in any verified propagation path to the observed symptom.
More broadly, step-wise causal ground truth raises the standard from correct localization to correct reasoning, the bar any diagnostic agent must clear to be trusted where errors carry real consequences.


\bibliography{ref}
\bibliographystyle{unsrtnat}

\newpage
\appendix
%

\noindent\textbf{Appendix organization.}
This appendix is organized into five parts.
\emph{Part~I} (Section~\ref{sec:appendix_datasheet}) is the dataset-side documentation: Datasheet, including a reproducibility section.
\emph{Part~II} (Sections~\ref{sec:appendix_systems}--\ref{sec:appendix_taxonomy}) specifies the benchmark testbed.
\emph{Part~III} (Section~\ref{sec:appendix_pipeline_details}) expands the \frameworkName{} pipeline compressed in Section~\ref{sec:methodology}.
\emph{Part~IV} (Section~\ref{sec:appendix_gt_validation}) reports the two-annotator audit on a random 100-case subsample and documents the calibration split used by the per-node anomaly screen.
\emph{Part~V} (Sections~\ref{sec:appendix_baselines}--\ref{sec:appendix_metrics}) documents the evaluated RCA agents and the formal metric definitions.

\section{Datasheet for Datasets}
\label{sec:appendix_datasheet}

\subsection{Motivation}
\datasetName{} is designed for \emph{process-level} evaluation of LLM-based RCA agents. The goal is to assess not only whether an agent identifies the correct root cause, but also whether it can justify that diagnosis with a faithful step-wise causal propagation path. Existing RCA benchmarks typically supervise only the final outcome label, which hides the \emph{ungrounded diagnosis} failure mode studied in this paper. No dedicated external funding was used specifically to compile the released benchmark artifact beyond the broader research effort supporting this study.

\subsection{Composition}
Each instance contains three components: (i)~multimodal pre- and post-injection telemetry, including distributed traces, metric time series, and structured logs in parquet format; (ii)~a structured ground-truth label specifying the originating system, the injected component, and the fault type; and (iii)~a step-wise causal propagation graph over services, pods, and spans (Section~\ref{sec:methodology}). \datasetName{} contains \totalValidLite{} evaluable instances drawn from \systemCount{} architecturally distinct microservice systems and covering \faultTypeCount{} chaos primary kinds; per-system breakdowns are reported in Table~\ref{tab:dataset_summary}. Curation uses two-pass stratified selection on (system, chaos family, root service) with per-stratum caps, sorted by (longest path, edge count, services), and retains only instances that pass end-to-end causal-path verification while excluding cyclic service graphs, trivial propagation depth ($\leq 1$), and frontend/loadgen entry points. The result is a curated, self-contained evaluation set whose sampling objective is balanced coverage rather than statistical representativeness of production incidents.

The current release does not define train/development/test splits. Retained instances are distributed as complete telemetry packs rather than field-pruned records, so field-level missingness or redaction is not a concern within the released trace, metric, or log data. Because the benchmark consists of synthetic operational telemetry generated under controlled load, it contains no PII or human-subject data; people-related subpopulation, identity, and sensitive-attribute considerations are therefore not applicable.

\subsection{Collection Process}
Faults are injected on three microservice systems: TrainTicket~\cite{zhou2018fault}, the OpenTelemetry Demo application~\cite{OpenTelemetryDemo}, and Hotel Reservation from DeathStarBench~\cite{DeathStarBench}. These interventions are run with Chaos Mesh on a shared 7-node Kubernetes cluster; per-system topology and deployment details are given in Section~\ref{sec:appendix_systems}, and the cluster configuration in Section~\ref{sec:appendix_cluster}. For each intervention, a fresh system instance is deployed and warmed up, a pre-injection baseline window is recorded, the fault is applied over $[t_0, t_0{+}\Delta t]$, and post-injection telemetry is then collected. Step-wise causal labels are produced automatically by the \frameworkName{} pipeline (Sections~\ref{sec:phase1_pruning}, \ref{sec:phase2_verification}) and are further checked by the forward verification protocol described later in this appendix.

Reliability of this procedure was assessed by a two-annotator audit on a uniform random 100-case subsample, with $94\%$ case-level agreement (Section~\ref{sec:exp_gt_validation}; full protocol in Appendix~\ref{sec:appendix_gt_validation}). The annotators were two of the paper's authors; no crowdworkers or paid external annotators were used. Because the benchmark is built from synthetic system telemetry rather than data about natural persons, consent, participant notification, and formal human-subject review considerations do not apply.

\subsection{Preprocessing, Cleaning, Labeling}
Telemetry is exported through OpenTelemetry into standardized parquet schemas, with one file per modality per period. The released benchmark contains only injections that pass the joint screen described in Section~\ref{sec:phase2_verification} (observable SLO impact, a rule-admitted propagation path, and a clean pre-injection baseline); cases failing any condition are dropped during pipeline-level analysis. The released package bundles the evaluation telemetry together with the derived labels and propagation graphs, so the released artifact is the processed benchmark itself rather than an undocumented intermediate product.

Cleaning is performed primarily at the instance level by filtering out runs whose pre-injection baseline is itself anomalous, rather than by manually editing fields inside retained telemetry records. The software responsible for preprocessing, filtering, and label construction is the released \frameworkName{} pipeline described in Section~\ref{sec:appendix_pipeline_details}. Pipeline reliability rests on the conjunctive verification design analyzed in Section~\ref{sec:appendix_thresholds} and is corroborated by the two-annotator audit reported in Section~\ref{sec:exp_gt_validation} (Appendix~\ref{sec:appendix_audit}).

\subsection{Uses}
Intended uses include (i)~process-level evaluation of LLM-based RCA agents and (ii)~training process reward models in the reliability domain, where the step-wise edges provide natural supervision targets. At the time of submission, the benchmark is used only in this study as the primary evaluation suite for process-level RCA assessment.

The benchmark is not intended for production-readiness certification of any specific RCA agent, for cross-architecture generalization claims to service-mesh, serverless, or fully event-driven systems beyond the three included testbeds, or for claims about faults that leave no telemetry signature. It should also not be used as a training corpus for the \emph{evaluated} RCA agents themselves, as doing so would create direct test-set leakage.

\subsection{Distribution and License}
An anonymized preview of the benchmark accompanies the supplementary submission, and the anonymized full benchmark package is provided in the submission materials. Upon acceptance, the full benchmark artifact and the \frameworkName{} pipeline will be released publicly with archival metadata and integrity checksums. The released benchmark is a derived artifact constructed from a curated subset of the upstream fault-injection runs together with the additional annotations and causal-process labels introduced in this work. The dataset is released under \textbf{CC-BY-SA~4.0}, and the pipeline code under \textbf{Apache~2.0}. To the best of our knowledge, no export-control or comparable regulatory restrictions apply to the released artifact.

\subsection{Maintenance}
To preserve anonymity in the submission version, direct maintainer contact details are omitted here and will be restored upon acceptance. Corrections, additional fault types, and cross-system extensions will be managed through versioned releases, and older public versions will remain archived for reproducibility. After public release, questions and fixes can be tracked through the GitHub issue tracker and pull requests. External contributions will be reviewed for schema compatibility, reproducibility, and annotation consistency before inclusion in an official release.

\subsection{Reproducibility}
\label{sec:appendix_repro}

The complete \frameworkName{} pipeline source is released alongside the dataset, with explicit random seeds for every stochastic component (DFS tie-breaking, baseline-window selection, fault-injection scheduling) so the numbers in this paper are regeneratable; pipeline hyperparameters and detector thresholds are tabulated in Sections~\ref{sec:appendix_hyperparams} and~\ref{sec:appendix_thresholds}, and cluster configuration is in Section~\ref{sec:appendix_cluster}.
A single-entry-point \texttt{reproduce.sh} accompanies the anonymized submission artifact and regenerates the Section~\ref{sec:exp_benchmark} numbers on a downloadable instance subset; full-dataset re-evaluation requires API quota for the \modelCount{} models documented in Section~\ref{sec:appendix_model_config}.
No pre-trained models are released as part of the benchmark.

\section{Benchmark Systems and Topology}
\label{sec:appendix_systems}

This section details the microservice systems used in our benchmark, including their architecture, scale, and the service dependency topology that defines the structural prior $\mathcal{G}$ for causal path extraction.

\subsection{Systems Overview}
\label{sec:appendix_system_overview}

The benchmark spans \systemCount{} open-source microservice testbeds chosen to cover distinct propagation regimes: TrainTicket~\cite{zhou2018fault} for deep, layered Java/REST call chains; the OpenTelemetry Demo application~\cite{OpenTelemetryDemo} for polyglot gRPC services with asynchronous Kafka legs; and Hotel Reservation from DeathStarBench~\cite{DeathStarBench} for shallow, wide Go/gRPC services with shared in-memory caches.
Together they cover three communication styles (synchronous REST, synchronous gRPC, and asynchronous queueing) and four data-store classes (relational, document, key-value cache, and message log), so the recovered ground-truth paths reflect propagation patterns that no single system would expose on its own.
The subsections below summarize the topology of each system; deployment manifests, image tags, and commit hashes are released alongside the dataset.

\subsection{TrainTicket Service Dependency Topology}
\label{sec:appendix_trainticket}

TrainTicket's 44 microservices follow a layered architecture centered on REST-style service-to-service calls, with RabbitMQ used for asynchronous messaging and a shared MySQL instance supporting many backend services. At a coarse level, requests flow from user-facing entry services through gateway and orchestration layers into domain services and then into shared infrastructure, creating long dependency chains and frequent opportunities for multi-hop propagation.

The system exhibits several properties that make RCA challenging:
\begin{itemize}[nosep, leftmargin=1.5em]
    \item \textbf{Deep call chains}: Booking operations traverse up to 6 service layers (Entry $\to$ Gateway $\to$ Orchestration $\to$ Domain $\to$ Foundation $\to$ Infrastructure).
    \item \textbf{High fan-in}: Foundation services (\texttt{Station}, \texttt{Route}, \texttt{Train}) are shared by many upstream callers, creating complex dependency patterns where a single failure can manifest in multiple symptom locations.
    \item \textbf{Shared infrastructure}: 23 services connect to a shared MySQL 5.7 instance, introducing vertical propagation paths from database-level faults to application-level errors.
    \item \textbf{Circular dependencies}: Some services exhibit mutual dependencies (e.g., \texttt{Travel} $\leftrightarrow$ \texttt{Seat}) that require temporal unrolling for causal analysis.
\end{itemize}

\subsection{OpenTelemetry Demo Service Dependency Topology}
\label{sec:appendix_otel_demo}

The OpenTelemetry Demo is a polyglot e-commerce reference application comprising 15 application services implemented across Go, Python, Java, .NET, Ruby, and JavaScript. The deployed stack also includes supporting components such as Kafka, Postgres, valkey/Redis, and observability infrastructure, which are not counted as application services.
Inter-service communication uses gRPC for synchronous request paths (e.g., \texttt{cart}~$\to$~\texttt{checkout}~$\to$~\texttt{payment}) and Kafka for asynchronous fan-out (e.g., \texttt{checkout}~$\to$~\texttt{accounting}~/~\texttt{fraud-detection}); state is held in Postgres, valkey/Redis, and the Kafka log itself.

Properties relevant for fault propagation:
\begin{itemize}[nosep, leftmargin=1.5em]
    \item \textbf{Polyglot error semantics}: each language stack surfaces failures differently (Java stack traces, Go panics, Python tracebacks, .NET exceptions), exercising the rule set across heterogeneous log formats.
    \item \textbf{Synchronous + asynchronous mix}: faults injected upstream of the Kafka boundary propagate as queue back-pressure rather than RPC timeouts, producing a propagation signature absent from the REST-only TrainTicket.
    \item \textbf{Shared infrastructure}: Postgres and valkey are accessed by multiple services; an injected database fault produces correlated anomalies across otherwise unrelated request paths.
\end{itemize}

\subsection{Hotel Reservation Service Dependency Topology}
\label{sec:appendix_hotelres}

Hotel Reservation, drawn from DeathStarBench~\cite{DeathStarBench}, consists of 9 Go services that communicate via gRPC and rely on MongoDB and Memcached.

Properties relevant for fault propagation:
\begin{itemize}[nosep, leftmargin=1.5em]
    \item \textbf{Shallow but wide call chains}: the maximum service-to-service depth is 3 hops (\texttt{frontend}~$\to$~\texttt{search}~$\to$~\texttt{profile}/\texttt{geo}/\texttt{rate}), with high fan-out at the search service.
    \item \textbf{In-memory cache layer}: Memcached sits between every read-heavy service and its MongoDB backing store; cache faults degrade many downstream services in parallel.
    \item \textbf{Terse error surface}: Go services return gRPC status codes directly rather than wrapping errors in stack traces, so log signals are sparser and trace-side evidence carries a comparatively larger share of the verification load.
\end{itemize}

\subsection{Propagation Channels and Telemetry Modalities}
\label{sec:appendix_channels_modalities}

The dependency graph $\mathcal{G}$ supports two propagation channels that the verification pipeline must distinguish.
\textit{Horizontal propagation} follows RPC call chains: a slow \texttt{Order} service causes its upstream callers to time out, producing cascading latency spikes along the call graph.
\textit{Vertical propagation} crosses abstraction layers: CPU throttling on a Kubernetes node degrades all co-located services simultaneously, even those with no mutual RPC dependency.
A single injection can trigger both channels, creating multi-hop paths that mix horizontal and vertical edges.

These channels leave distinct signatures in the three telemetry modalities, which Phase~2 verification (Section~\ref{sec:phase2_verification}) exploits.
\textit{Traces} record per-request call trees with span-level latencies and status codes; horizontal propagation manifests as anomalous parent$\to$child span timing that reveals the cascade direction.
\textit{Metrics} are per-service and per-host time series (CPU usage, p99 latency, error rate); vertical propagation appears as correlated anomalies among co-located services sharing a throttled resource.
\textit{Logs} are semi-structured text events (e.g., \texttt{[ERROR] upstream timeout}) that supply causal context absent from numerical signals alone.

\section{Root Cause Taxonomy}
\label{sec:appendix_taxonomy}

This section formalizes the ground truth label schema and provides the complete fault type catalog.

\subsection{Ground Truth Schema}
\label{sec:appendix_gt_schema}

Each fault injection instance is annotated with a structured ground truth label that specifies the root cause at multiple granularity levels. 
Table~\ref{tab:gt_schema} shows the schema fields; cardinalities vary by system, and TrainTicket values are shown as concrete examples.

\begin{table}[t]
\centering
\caption{Ground truth label schema. Each fault injection instance is annotated with a structured label across multiple granularity levels. Cardinalities are system-dependent; example values are drawn from TrainTicket for concreteness.}
\label{tab:gt_schema}
\resizebox{\columnwidth}{!}{%
\begin{tabular}{llcl}
\toprule
Field & Description & Cardinality & Example Value \\
\midrule
\texttt{service} & Target microservice & system-dependent & \texttt{ts-order-service} \\
\texttt{pod} & Target Kubernetes Pod & system-dependent & \texttt{ts-order-service-7b8f4-xk9z2} \\
\texttt{container} & Target container within Pod & system-dependent & \texttt{ts-order-service} \\
\texttt{function} & Target function or handler (code-level faults) & system-dependent & \texttt{com.order.service.impl.create} \\
\texttt{span} & Target trace span (HTTP faults) & system-dependent & \texttt{POST /api/v1/orderservice/order} \\
\texttt{metric} & Anomalous golden signal & system-dependent & \texttt{http.server.request.duration} \\
\bottomrule
\end{tabular}%
}
\end{table}

The multi-level annotation enables evaluation at different abstraction levels. 
Service-level evaluation ($|\mathcal{V}| = 44$) is the most commonly reported granularity in prior work. 
Finer-grained levels (function, span) provide more precise localization but increase label space complexity substantially.

\subsection{Complete Fault Type Catalog}
\label{sec:appendix_fault_types}

Table~\ref{tab:fault_taxonomy} presents the fault types implemented in our chaos engineering framework, organized by Chaos Mesh CRD category.
Each fault type is classified by:
\begin{itemize}[nosep, leftmargin=1.5em]
    \item \textbf{Target layer}: Whether the fault targets infrastructure resources (Pod, network, CPU/memory) or application logic (HTTP endpoints, JVM methods, database queries).
    \item \textbf{Applicable ground truth levels}: Which fields of the ground truth schema are relevant. Infrastructure faults are labeled at the \texttt{service}/\texttt{pod} level, while application faults include \texttt{function} or \texttt{span} annotations.
    \item \textbf{Propagation channel}: Whether the fault propagates horizontally (between logical entities via RPC) or vertically (from infrastructure to logical layers via resource contention), or both.
\end{itemize}

\begin{table}[t]
\centering
\caption{Fault type taxonomy across 7 Chaos Mesh CRD categories. ``Applicable Levels'' lists the ground-truth fields relevant to each fault type.}
\label{tab:fault_taxonomy}
\resizebox{\columnwidth}{!}{%
\begin{tabular}{llllll}
\toprule
Category & Fault Type & Parameters & Target Layer & Applicable Levels & Propagation Channel \\
\midrule
\multirow{3}{*}{PodChaos}
  & PodKill & Kill signal (SIGKILL) & Infrastructure & service, pod & Vertical \\
  & PodFailure & Unavailability duration & Infrastructure & service, pod & Vertical \\
  & ContainerKill & Target container & Infrastructure & service, pod, container & Vertical \\
\midrule
\multirow{2}{*}{StressChaos}
  & CPUStress & Core count, load \% & Infrastructure & service, pod & Vertical \\
  & MemoryStress & Memory size (MB) & Infrastructure & service, pod & Vertical \\
\midrule
\multirow{9}{*}{HTTPChaos}
  & HTTPRequestAbort & Abort code, target path & Application & service, span & Horizontal \\
  & HTTPResponseAbort & Abort code, target path & Application & service, span & Horizontal \\
  & HTTPRequestDelay & Delay (ms), target path & Application & service, span & Horizontal \\
  & HTTPResponseDelay & Delay (ms), target path & Application & service, span & Horizontal \\
  & HTTPResponseReplaceBody & Body type \{empty, random\} & Application & service, span & Horizontal \\
  & HTTPResponsePatchBody & Patch content & Application & service, span & Horizontal \\
  & HTTPRequestReplacePath & New path & Application & service, span & Horizontal \\
  & HTTPRequestReplaceMethod & Method \{GET, POST, ...\} & Application & service, span & Horizontal \\
  & HTTPResponseReplaceCode & Status code & Application & service, span & Horizontal \\
\midrule
\multirow{2}{*}{DNSChaos}
  & DNSError & Target hostname & Infrastructure & service, pod & Vertical \\
  & DNSRandom & Target hostname & Infrastructure & service, pod & Vertical \\
\midrule
\multirow{6}{*}{NetworkChaos}
  & NetworkDelay & Delay (ms), jitter & Infrastructure & service, pod & Vertical + Horizontal \\
  & NetworkLoss & Loss rate (\%) & Infrastructure & service, pod & Vertical + Horizontal \\
  & NetworkDuplicate & Duplicate rate (\%) & Infrastructure & service, pod & Vertical + Horizontal \\
  & NetworkCorrupt & Corrupt rate (\%) & Infrastructure & service, pod & Vertical + Horizontal \\
  & NetworkBandwidth & Rate (bps) & Infrastructure & service, pod & Vertical + Horizontal \\
  & NetworkPartition & Partition direction & Infrastructure & service, pod & Vertical + Horizontal \\
\midrule
\multirow{1}{*}{TimeChaos}
  & TimeSkew & Offset (ms) & Infrastructure & service, pod & Vertical \\
\midrule
\multirow{8}{*}{JVMChaos}
  & JVMLatency & Delay (ms), target class.method & Application & service, function & Horizontal \\
  & JVMReturn & Return value, type \{String, Int\} & Application & service, function & Horizontal \\
  & JVMException & Exception class, target method & Application & service, function & Horizontal \\
  & JVMGarbageCollector & --- & Application & service, pod & Vertical \\
  & JVMCPUStress & CPU count & Application & service, pod & Vertical \\
  & JVMMemoryStress & Type \{heap, stack\}, size & Application & service, pod & Vertical \\
  & JVMMySQLLatency & Delay (ms), target method & Application & service, function & Horizontal \\
  & JVMMySQLException & Exception, target method & Application & service, function & Horizontal \\
\bottomrule
\end{tabular}%
}
\end{table}

\subsection{Observable Fault-Kind Vocabulary for Evaluation}
\label{sec:appendix_kind_vocab}

\begin{table}[t]
\centering
\caption{Mapping from raw \texttt{chaos\_type} (Table~\ref{tab:fault_taxonomy}) to the observable fault-kind labels used at evaluation time. Rows listing multiple source identifiers are many-to-one collapses where the listed mechanisms produce indistinguishable telemetry signatures; the remaining rows map one-to-one.}
\label{tab:kind_vocabulary}
\resizebox{\columnwidth}{!}{%
\begin{tabular}{@{}lll@{}}
\toprule
Fault kind & Source \texttt{chaos\_type}(s) & Primary observation channel \\
\midrule
\multicolumn{3}{@{}l}{\emph{Pod / container lifecycle}} \\
\texttt{pod\_failure}                    & \texttt{PodKill}, \texttt{ContainerKill}        & trace emission gap, brief, recovers in window \\
\texttt{pod\_unavailable}                & \texttt{PodFailure}                             & trace emission gap, sustained to window end \\
\midrule
\multicolumn{3}{@{}l}{\emph{L3/L4 network (direction recorded in GT)}} \\
\texttt{network\_delay}                  & \texttt{NetworkDelay}                           & \texttt{container.network.*}, trace latency tail \\
\texttt{network\_loss}                   & \texttt{NetworkLoss}                            & \texttt{container.network.*}, trace retry pattern \\
\texttt{network\_partition}              & \texttt{NetworkPartition}                       & \texttt{container.network.*}, timeout share approaches one \\
\texttt{network\_corrupt}                & \texttt{NetworkCorrupt}                         & \texttt{container.network.*}, TCP retransmit signal \\
\texttt{network\_duplicate}              & \texttt{NetworkDuplicate}                       & \texttt{container.network.*}, duplicate-ack pattern \\
\texttt{network\_bandwidth\_limit}       & \texttt{NetworkBandwidth}                       & \texttt{container.network.*}, throughput drop \\
\midrule
\multicolumn{3}{@{}l}{\emph{HTTP application layer}} \\
\texttt{http\_aborted}                   & \texttt{HTTPRequestAbort}, \texttt{HTTPResponseAbort} & HTTP span status, request-failure rate \\
\texttt{http\_slow}                      & \texttt{HTTPRequestDelay}, \texttt{HTTPResponseDelay} & HTTP span duration distribution \\
\texttt{http\_payload\_modified}         & \texttt{HTTPResponseReplaceBody}, \texttt{HTTPResponsePatchBody}, \texttt{HTTPRequestReplacePath}, \texttt{HTTPRequestReplaceMethod} & response body / path / method differs from baseline \\
\texttt{http\_response\_status\_modified}& \texttt{HTTPResponseReplaceCode}                & HTTP span status-code distribution \\
\midrule
\multicolumn{3}{@{}l}{\emph{Resource exhaustion}} \\
\texttt{cpu\_stress}                     & \texttt{CPUStress}                              & \texttt{container.cpu.*} saturated \\
\texttt{mem\_stress}                     & \texttt{MemoryStress}                           & \texttt{container.memory.*} saturated \\
\texttt{jvm\_thread\_cpu\_stress}        & \texttt{JVMCPUStress}                           & \texttt{jvm.cpu.*} elevated, container CPU normal \\
\texttt{jvm\_heap\_stress}               & \texttt{JVMMemoryStress}                        & \texttt{jvm.memory.used} climbs to limit \\
\texttt{jvm\_gc\_pressure}               & \texttt{JVMGarbageCollector}                    & \texttt{jvm.gc.duration} histogram spikes \\
\midrule
\multicolumn{3}{@{}l}{\emph{Code-level JVM}} \\
\texttt{jvm\_method\_exception}          & \texttt{JVMException}                           & log exception class on a specific method \\
\texttt{jvm\_jdbc\_exception}            & \texttt{JVMMySQLException}                      & exception on a JDBC client call \\
\texttt{jvm\_method\_latency}            & \texttt{JVMLatency}                             & elevated span duration on a method \\
\texttt{jvm\_jdbc\_latency}              & \texttt{JVMMySQLLatency}                        & elevated span duration on a JDBC call \\
\texttt{jvm\_method\_mutated}            & \texttt{JVMReturn}, \texttt{JVMRuntimeMutator}  & method returns a value differing from baseline \\
\midrule
\multicolumn{3}{@{}l}{\emph{DNS / clock}} \\
\texttt{dns\_resolution\_failed}         & \texttt{DNSError}, \texttt{DNSChaos}            & DNS error log entries on the affected service \\
\texttt{dns\_resolution\_wrong}          & \texttt{DNSRandom}                              & resolution succeeds but returns wrong target \\
\texttt{clock\_skew}                     & \texttt{TimeSkew}, \texttt{TimeChaos}           & cross-service timestamp drift in correlated logs \\
\bottomrule
\end{tabular}%
}
\end{table}

The chaos-mesh API names in Table~\ref{tab:fault_taxonomy} describe \emph{how} a fault is injected, not what an agent observes from telemetry; in many cases distinct injection mechanisms produce indistinguishable signatures in the parquet data (for example \texttt{HTTPRequestAbort} and \texttt{HTTPResponseAbort} both surface as request failures from the client side, and \texttt{PodKill} and \texttt{ContainerKill} both manifest as a brief gap in the affected service's trace emission).
An agent therefore cannot recover the raw chaos-mesh API name from the released telemetry alone.
We define an observable fault-kind vocabulary by relabeling each chaos mechanism according to the feature signature it produces in metrics, traces, and logs, and judge an agent correct when its predicted label matches this observable label rather than the underlying API name.
Table~\ref{tab:kind_vocabulary} lists the mapping from raw \texttt{chaos\_type} to observable label together with the primary signal carrying each label; the same projection is applied at ground-truth extraction and at agent-prediction matching so both sides of the comparison share one vocabulary.

\section{\frameworkName{} Pipeline and Experimental Details}
\label{sec:appendix_pipeline_details}

This appendix details (i)~the pipeline internals compressed in Section~\ref{sec:methodology} for space (Sections~\ref{sec:appendix_scm}--\ref{sec:appendix_time_expanded_graph}, plus the rule set, hyperparameters, and detector thresholds in Sections~\ref{sec:appendix_rules}, \ref{sec:appendix_hyperparams}, \ref{sec:appendix_thresholds}), and (ii)~the operational and calibration details supporting reproduction (Sections~\ref{sec:appendix_cluster}--\ref{sec:appendix_thresholds}).

\subsection{Structural Causal Model}
\label{sec:appendix_scm}

Fault propagation over $\mathcal{G}$~\cite{pearl2009causality} is realized at each node as
\begin{equation}
    S_i^{(t)} = f_i\!\bigl(\mathbf{S}_{Pa(v_i)}^{(<t)},\; U_i^{(t)}\bigr),
    \label{eq:scm}
\end{equation}
where $\mathbf{S}_{Pa(v_i)}^{(<t)}$ are the recent states of $v_i$'s causal parents and $U_i^{(t)}$ is an exogenous shock.
The agent's observation $\mathbf{O}_{obs} \subset \mathbf{O}_{true}$ is a strict subset of system state (instrumentation gaps, sampling), which is the source of the information asymmetry in inequality~(\ref{eq:info_asymmetry}).

\subsection{Phase 1: Discrete State Abstraction \texorpdfstring{$\Psi$}{Psi}}
\label{sec:appendix_state_abstraction}

We project high-dimensional raw telemetry $\mathbf{O}^{(t)} \in \mathbb{R}^d$ onto a discrete state alphabet $\Sigma$ via a node-type-aware mapping $\Psi(O_i, \theta(v_i)) \to s_i \in \Sigma$, where $\theta(v_i)$ is the type of node $v_i$ (matching the type tag $\theta_i$ used in the Phase~1 rule transitions of Section~\ref{sec:phase1_pruning}). For \emph{logical nodes} (services, spans), states capture performance violations such as \texttt{HighLatency} and \texttt{HighErrorRate}; for \emph{infrastructure nodes} (pods, containers, hosts), states reflect resource exhaustion such as \texttt{CPU\_Throttled} and \texttt{MemoryPressure}.
This abstraction filters high-frequency noise ($U_i$), retaining only semantically meaningful events as inputs to the rule-filtering and DFS stages.

\subsection{Phase 1: Time-Expanded Heterogeneous Graph and Constrained DFS}
\label{sec:appendix_time_expanded_graph}

To enumerate candidate paths $\Pi_{cand}$, we construct a Time-Expanded Heterogeneous Graph whose nodes are tuples $(v, s, t)$: a system node $v\in\mathcal{V}$ with state $s\in\Sigma$ activated at time $t$. Edges connect tuples whose state transitions are admitted by the propagation rule set $\mathcal{R}$ (Section~\ref{sec:appendix_rules}) and whose times are causally ordered. A constrained Depth-First Search starting from the injection point unrolls circular dependencies (common in microservice meshes) into acyclic chains, with the configured maximum path length and per-node revisit budget (Section~\ref{sec:appendix_hyperparams}). The resulting paths are topologically valid sequences of state transitions; statistical and temporal verification (Section~\ref{sec:phase2_verification}) then admits or rejects them.

\subsection{Cluster Configuration}
\label{sec:appendix_cluster}

Our testbed runs on Kubernetes v1.29 with the following specifications:
\begin{itemize}[nosep, leftmargin=1.5em]
    \item \textbf{Cluster}: 1 control-plane node (\texttt{master1}, 48 vCPU, 64GB RAM) and 6 worker nodes running Debian 12 with containerd 1.7.5; \texttt{worker1}--\texttt{worker3} each provide 32 vCPU and 64GB RAM, while \texttt{worker4}--\texttt{worker6} each provide 128 vCPU and 128GB RAM
    \item \textbf{Systems hosted}: TrainTicket (44 Java/Spring services), OpenTelemetry Demo (15 polyglot services), and Hotel Reservation (9 Go services), each deployed in separate Kubernetes namespaces on the shared cluster
    \item \textbf{Service-count convention}: Reported per-system service counts are application \texttt{Deployments} obtained via \texttt{kubectl get deploy} on the running benchmark, excluding load generators, the OpenTelemetry collector, datastores, message brokers, and the TrainTicket UI dashboard
    \item \textbf{Deployment versions}: Exact system commit hashes and image tags for the three deployments are pinned in the archived manifests released with the benchmark
    \item \textbf{Traffic load}: Synthetic user sessions are generated by per-system load generators after a warm-up period, with sustained request streams used to establish pre-injection baselines and expose post-injection propagation
    \item \textbf{Monitoring stack}: OpenTelemetry-based collection with per-system telemetry export to standardized parquet artifacts, together with cluster-level collectors and Prometheus-based metric scraping
    \item \textbf{Fault injection}: Chaos Mesh with programmatic CRD generation for application, resource, network, and pod-level interventions
    \item \textbf{Stateful backends}: Across the three systems, the benchmark exercises MySQL, PostgreSQL, MongoDB, Memcached/valkey, RabbitMQ, and Kafka-backed propagation paths
\end{itemize}

\subsection{\frameworkName{} Hyperparameters}
\label{sec:appendix_hyperparams}

Key parameters for the \frameworkName{} annotation pipeline:
\begin{itemize}[nosep, leftmargin=1.5em]
    \item \textbf{State Detection Window $\Delta$}: 3\,s for span/service-level metrics; 5\,s for pod, container, and machine-level metrics
    \item \textbf{State Activation}: Dual detection combines fixed utilization thresholds ({$>$}80\% CPU/memory) with Z-score baseline comparison ($Z > 3\sigma$ for critical anomalies), plus adaptive multiplier thresholds for latency metrics based on coefficient of variation (CV)
    \item \textbf{Maximum Path Length}: 5 hops (default), with diamond-shaped revisits allowed (max 2 visits per node)
    \item \textbf{Rule Confidence}: Default 0.8 per propagation rule, with per-rule overrides
\end{itemize}

\subsection{Propagation Rule Set}
\label{sec:appendix_rules}
\begin{table}[t]
\centering
\caption{Complete propagation rule set $\mathcal{R}$. \textit{Src}/\textit{Dst} denote source and destination entity types; \textit{Path} shows intermediate hops for multi-hop rules.}
\label{tab:propagation_rules}
\resizebox{\columnwidth}{!}{%
\begin{tabular}{lllllll}
\toprule
ID & Category & Src Kind : States & Path & Dst Kind : States & Mechanism \\
\midrule
\multicolumn{6}{l}{\textit{Vertical: Container $\to$ Span}} \\
C-01 & vertical & container : \{cpu, mem\} & pod $\to$ svc $\to$ & span : \{lat, p99, timeout, missing, err\} & Resource contention \\
C-02 & vertical & container : \{restart\} & pod $\to$ svc $\to$ & span : \{missing, err, timeout\} & Process termination \\
\midrule
\multicolumn{6}{l}{\textit{Vertical: Pod $\to$ Span}} \\
P-01 & vertical & pod : \{cpu, mem\} & svc $\to$ & span : \{lat, p99, timeout\} & Resource contention \\
\midrule
\multicolumn{6}{l}{\textit{First-hop: Service $\to$ Span}} \\
S-01 & first-hop & service : \{any\} & --- & span : \{lat, p99, timeout\} & Latency inheritance \\
S-02 & first-hop & service : \{err, unavail\} & --- & span : \{err, timeout, reset, missing, log\_err\} & Error inheritance \\
\midrule
\multicolumn{6}{l}{\textit{Horizontal: Span $\to$ Span (RPC cascade)}} \\
H-01 & horizontal & span : \{lat, p99\} & --- & span : \{lat, p99, timeout\} & Callee slowdown \\
H-02 & horizontal & span : \{err, timeout\} & --- & span : \{err, timeout\} & Error propagation \\
H-03 & horizontal & span : \{missing\} & --- & span : \{err, timeout, missing\} & Missing callee \\
H-04 & horizontal & span : \{err, missing\} & healthy span $\to$ & span : \{err, missing\} & Controller bypass \\
H-05 & horizontal & span : \{err, timeout\} & --- & span : \{timeout\} & Retry-induced wait \\
H-06 & horizontal & span : \{inj\_affected\} & --- & span : \{err, lat, p99, timeout, missing\} & Infra fault relay \\
H-07 & horizontal & span : \{healthy\} & healthy span $\to$ & span : \{err, lat, p99, missing\} & JVM injection bypass \\
\midrule
\multicolumn{6}{l}{\textit{Cross-channel: Span $\to$ Pod $\to$ Span}} \\
X-01 & cross & span : \{lat, p99\} & svc $\to$ pod $\to$ svc $\to$ & span : \{lat, p99, timeout\} & JVM stress via shared pod \\
\bottomrule
\end{tabular}%
}
\end{table}

Table~\ref{tab:propagation_rules} lists the complete rule set $\mathcal{R}$ used in Phase~1 (Section~\ref{sec:phase1_pruning}).
Rules are constructed by enumerating all (source entity type $\times$ source state $\times$ edge type $\times$ destination entity type) combinations in the system dependency graph, then retaining only those for which a known physical or logical mechanism exists.
We group rules into four categories by injection granularity.

\textbf{Vertical rules} (container$\to$span, pod$\to$span) encode resource contention: a throttled container or pod degrades all spans of its co-located service through the Kubernetes scheduling hierarchy (Container $\subset$ Pod $\subset$ Node).
\textbf{First-hop rules} (service$\to$span) bridge the aggregation layer, since services are logical groupings without directly observable states.
\textbf{Horizontal rules} (span$\to$span) capture RPC call-chain cascades, the dominant propagation channel, accounting for over 80\% of observed edges.
\textbf{Cross-channel rules} (span$\to$pod$\to$span) model JVM stress scenarios where resource contention on a shared pod affects sibling services.

The rule set is intentionally over-inclusive: rules with zero empirical usage (e.g., service error$\to$span) are retained for completeness, and Phase~2's statistical verification (Section~\ref{sec:phase2_verification}) filters false positives.
Rules lacking any empirical support after 500+ injection experiments were removed in version 2.0 (e.g., circuit-breaker bypass, pod-level disk/network faults), as noted in the version history.

\subsection{Scalability and Computational Cost}
\label{sec:appendix_scalability}

All experiments were conducted on a virtual machine with 32 CPU cores and 64GB RAM.
The path extraction process for a single fault injection scenario, covering a 15-minute telemetry window ($\sim$1GB of metrics and traces), completes in an average of 3.5 minutes.
Computational complexity scales with the number of active anomalous entities rather than total system components, demonstrating practical scalability.

\subsection{Anomaly Detection Thresholds}
\label{sec:appendix_thresholds}

The per-node anomaly screen in Phase~2 (Section~\ref{sec:phase2_verification}) uses three detector families, each matched to the distributional characteristics of the underlying metric.

\textbf{Z-score detector} (for approximately Gaussian metrics such as CPU and memory utilization).
A metric value $x$ is flagged if its Z-score $z = (x - \mu_{base}) / \max(\sigma_{base},\, \epsilon)$ exceeds a threshold, where $\mu_{base}$ and $\sigma_{base}$ are the baseline mean and standard deviation and $\epsilon = 10^{-6}$ prevents division by zero.
We use three severity levels: \textit{critical} ($z > 3.0$), \textit{warning} ($z > 2.0$), and \textit{moderate} ($z > 1.5$).

\textbf{Percentile detector} (for heavy-tailed metrics such as error counts).
Instead of parametric assumptions, anomalies are flagged relative to empirical baseline percentiles: \textit{critical} if the value exceeds the baseline P99.9 or $2{\times}$P99; \textit{warning} if it exceeds P99 or $1.5{\times}$P99; \textit{moderate} if it exceeds P90 or $1.2{\times}$P99.

\textbf{Adaptive latency detector} (for request latency, which spans orders of magnitude across services).
The threshold multiplier is $\tau = (\tau_{cv\text{-}base} + \tau_{cv\text{-}range} \cdot (1 - e^{-cv / \alpha})) \cdot (l_{ref} / l_{base})^{\beta}$, where $cv$ is the baseline coefficient of variation, $l_{base}$ the baseline mean latency, and $l_{ref}$ a reference latency.
This formulation encodes the principle that low-baseline services (e.g., 1\,ms mean) tolerate large relative spikes with minimal user impact, while high-baseline services (e.g., 5\,s mean) require tighter thresholds.

\textbf{Design rationale.}
All three detectors are intentionally permissive at the per-node level: they aim to avoid false negatives (missing a genuinely affected node) at the cost of admitting some false positives (flagging coincidental deviations).
False positives are subsequently eliminated by the conjunctive verification criterion: a candidate path is accepted only when every node passes the anomaly screen, every edge conforms to the propagation rule set $\mathcal{R}$, and downstream anomalies follow upstream causes in time.
This layered design decouples the sensitivity of individual detectors from the overall precision of the ground truth.

\textbf{Why conjunction filtering is effective.}
A candidate path $\pi$ of length $k$ is verified only if three independent conditions hold simultaneously at every hop: (i)~statistical anomaly, (ii)~structural rule conformance, and (iii)~temporal ordering.
Let $p_s$, $p_r$, and $p_t$ denote the per-hop false-positive rates of each condition in isolation.
Under an independence assumption, the probability that a spurious (non-causal) path of length $k$ accidentally satisfies all three conditions at every hop is bounded by:
\begin{equation}
    P(\text{false path accepted}) \;\leq\; (p_s \cdot p_r \cdot p_t)^{k}
    \label{eq:conjunction_fp}
\end{equation}
Even with individually permissive thresholds (e.g., $p_s{=}0.2$, $p_r{=}0.3$, $p_t{=}0.5$), the per-hop joint false-positive rate is $0.2 \times 0.3 \times 0.5 = 0.03$, and for a typical path of length $k{=}3$ this yields $0.03^3 \approx 2.7 \times 10^{-5}$.
This exponential decay in $k$ means that longer propagation paths, which are harder to verify, are also the most resistant to spurious acceptance.
The independence assumption is approximate: in practice the three conditions share some information (e.g., structural constraints partially determine temporal ordering).
Nevertheless, the exponential scaling ensures robust filtering even when the conditions are moderately correlated.

\section{Audit Protocol and Calibration}
\label{sec:appendix_gt_validation}

This appendix details the two-annotator audit conducted on a random 100-case subsample of \datasetName{} (Section~\ref{sec:exp_gt_validation}) and the calibration split used by the per-node anomaly screen.
Soundness of the verified causal subgraph $\mathcal{G}^*$ rests on the conjunctive verification design analyzed in Section~\ref{sec:appendix_thresholds}: $do(v_{root})$ pins the root direction, and an admitted path must clear three independently sourced gates simultaneously.
The audit reported here is the empirical complement to that design argument.

\subsection{Annotator Audit on a Random Subsample}
\label{sec:appendix_audit}

\textbf{Goal.}
Bound the reliability of $\mathcal{G}^*$ by measuring agreement between the verified causal subgraphs produced by \frameworkName{} and the propagation paths that domain-experienced annotators consider plausible from the same telemetry.

\textbf{Sampling.}
A simple random sample of $N{=}100$ instances was drawn without replacement from the \totalValidLite{} evaluable instances of \datasetName{}.
Sampling was uniform rather than stratum-balanced, so the audit's verdict reflects the dataset's natural mix of (system, fault family) pairs; post-hoc, the sample covered all \systemCount{} systems and every chaos primary kind retained in the benchmark.

\textbf{Annotators.}
Two of the paper's authors served as annotators.
Both have research experience in distributed-system observability and operational fault analysis, and are familiar with the \frameworkName{} pipeline output schema.
The annotators worked independently from disjoint copies of each case's telemetry and were blind to one another's verdicts until the adjudication step.

\textbf{Materials per case.}
For each instance, an annotator received: (i)~the multimodal telemetry over both pre- and post-injection windows (distributed traces, metric series, structured logs, all in their original parquet schema); (ii)~the injected-fault metadata, comprising $v_{root}$, fault type, and the injection window $[t_0, t_0{+}\Delta t]$; and (iii)~the pipeline-produced cascade $\mathcal{G}^*$ rendered as a labelled directed graph over services, pods, and spans.
The pipeline output was shown so the annotator could assess it directly; the underlying telemetry was provided so the assessment was grounded in the raw evidence rather than the pipeline's own intermediate features.

\textbf{Annotation rubric.}
Each annotator recorded a single binary verdict per case: \textsc{Agree} when every edge in $\mathcal{G}^*$ is plausibly causal given the telemetry \emph{and} no obvious propagation edge is missing; \textsc{Disagree} otherwise.
When an annotator marked \textsc{Disagree}, they were instructed to flag the specific edge or path that was either implausible (an included edge with no telemetry support) or missing (a propagation step visible in the telemetry but absent from $\mathcal{G}^*$).

\textbf{Verification procedure.}
The binary verdict is grounded in a fixed three-stage inspection of each case, applied uniformly across the sample.
The annotator first reviews the injected-fault metadata to anchor expectation: which component received the intervention $do(v_{root})$, which fault primitive was applied, and the injection window $[t_0, t_0{+}\Delta t]$.
The annotator then proceeds through three stages.

\emph{Stage 1: anomaly verification (per node).}
For each node in $\mathcal{G}^*$ that carries a non-healthy state annotation, the annotator opens the underlying telemetry slice for that node over the abnormal window (span series for span nodes, metric and log streams for service and pod nodes) and verifies that the annotated abnormality is visible in the raw evidence.
A node whose label cannot be located in the raw signal is flagged as a detector false positive.

\emph{Stage 2: hop-level causality verification (per edge).}
For each edge $u \to v$ in $\mathcal{G}^*$, the annotator assesses three properties: (i)~\emph{temporal precedence}, where $u$'s onset of abnormality precedes $v$'s within plausible propagation latency; (ii)~\emph{structural plausibility}, where the architectural relationship between $u$ and $v$ (caller--callee, container-of-pod, span-within-service, or other declared edge kind) admits the proposed propagation mechanism for the injected fault primitive; and (iii)~\emph{causal versus coincidental}, where the change in $v$ is consistent with a downstream consequence of the abnormality at $u$ (correlated latency, error-rate, or saturation pattern) rather than an unrelated fluctuation that happens to fall within the same time window.
An edge that fails any of (i)--(iii) is flagged as implausible.

\emph{Stage 3: coverage check.}
The annotator scans the telemetry for nodes that exhibit clear abnormality during the injection window but are absent from $\mathcal{G}^*$.
For each such omission the annotator asks whether the missing node lies on a plausible propagation hop from $v_{root}$ to a labelled alarm in $\mathcal{G}^*$; if it does, the missing path is flagged.

A case receives \textsc{Agree} only when stages~1--3 produce no flags; any flag triggers \textsc{Disagree}, and the flagged node, edge, or missing path is recorded for adjudication.

\textbf{Adjudication.}
After both annotators completed the sample, every case on which the two verdicts diverged was reviewed jointly.
The two annotators discussed the flagged edges, returned to the underlying telemetry where needed, and reached consensus on the source of each disagreement.
This adjudication produced a per-disagreement note used in the analysis below; it did not modify the released $\mathcal{G}^*$ for those cases, since the audit is a measurement of the pipeline's output, not a relabelling pass.

\textbf{Result.}
Inter-annotator agreement on the case-level verdict was $94/100 = 94\%$.
The six disagreements were entirely one-directional: in each case, at least one annotator considered an additional propagation path plausible that $\mathcal{G}^*$ did not include.
No case was flagged in the opposite direction, that is, no annotator considered any edge already in $\mathcal{G}^*$ implausible, so within the audited sample the released cascades contain no detected false-positive edges.

\textbf{Disagreement analysis.}
Joint review attributed every under-coverage instance to the conjunctive verification design's deliberate strictness on borderline evidence: the gates had rejected paths whose statistical-deviation, rule-conformance, or temporal-alignment signal sat near the admission threshold.
The flagged paths were not spurious (annotators considered them plausible) but they were weakly supported, and in each instance the path corresponded to a service receiving low traffic during the abnormal window, where the per-node anomaly screen could not cleanly separate the propagated effect from baseline jitter.
This direction of error is consistent with the protocol's stated property: $\mathcal{G}^*$ is constructed as a high-precision over-approximation of the underlying causal structure, prioritizing precision (no spurious edges admitted) over recall (some weakly supported edges may be missed).
For \datasetName{}'s intended use as a process-level reference for evaluating LLM agents' propagation-path reasoning, admitting a false edge into the reference would directly mis-credit or mis-penalize an agent's diagnosis; omitting a weakly supported edge does not.
The audit therefore confirms the design's labelling discipline: in the audited sample the released cascades are sound (no implausible included edges) at the cost of conservative coverage, which we accept as the appropriate trade-off for the benchmark's intended use.

\subsection{Baseline Cal-Eval Split}
\label{sec:appendix_cal_eval}

For each instance, the pre-injection baseline window $\mathcal{D}_{base}$ is split into two halves: the first calibrates per-node thresholds (Z-score parameters, percentile cutoffs, adaptive-latency multipliers; see Section~\ref{sec:appendix_thresholds}), the second evaluates them. Calibrating and evaluating on the same window inflates the per-node admission rate because the screen would fit the noise it then tests against; the split prevents this overfit. Default calibration set size is half the baseline window length ($\sim 7.5$ minutes).

\section{LLM-Based RCA Agent Details}
\label{sec:appendix_baselines}

This section details the evaluation framework, agent architecture, and prompt design for the LLM-based RCA agents evaluated in our benchmark.

\subsection{Agent Evaluation Framework}
\label{sec:appendix_framework}

Each LLM agent receives a datapack containing multimodal telemetry (traces, metrics, logs in parquet format) and produces a structured CausalGraph: a directed graph with annotated root causes, affected nodes, and causal edges.
Evaluation operates at two levels: outcome-level (root cause accuracy) and process-level (causal edge and node fidelity against ground truth).
This dual evaluation design enables the stratified Edge~F1 analysis presented in Section~\ref{sec:exp_benchmark}.

\subsection{Agent Architecture}
\label{sec:appendix_agent_arch}

All agents follow a tool-augmented LLM architecture adapted from the Deep Research agent~\cite{lin2025deepresearch}, operating in a ReAct-style~\cite{yao2022react} investigation loop with three nodes:

\begin{enumerate}[nosep, leftmargin=1.5em]
    \item \textbf{\texttt{llm\_call}} (Decision Node): The LLM analyzes the current investigation state, consisting of the incident description, prior tool observations, and its own reflections, and either issues additional tool calls or terminates the loop.
    \item \textbf{\texttt{tool\_node}} (Execution Node): Dispatches all tool calls from the LLM response, executes them asynchronously, and returns observations as \texttt{ToolMessage} objects. Errors are caught and returned as observations (the loop does not crash on tool failures).
    \item \textbf{\texttt{compress\_rca\_findings}} (Synthesis Node): Invoked when the LLM stops issuing tool calls. A separate LLM call synthesizes all accumulated evidence into a structured CausalGraph JSON.
\end{enumerate}

\noindent The control flow follows:
\[
    \texttt{START} \to \texttt{llm\_call} \xrightarrow{\text{tool\_calls?}} \texttt{tool\_node} \to \texttt{llm\_call} \to \cdots \to \texttt{compress} \to \texttt{END}
\]

\noindent\textbf{Tool suite.} The agent has access to four tools:
\begin{itemize}[nosep, leftmargin=1.5em]
    \item \texttt{list\_tables\_in\_directory}: Discovers available parquet files and their metadata (row/column counts).
    \item \texttt{get\_schema}: Inspects the column names, types, and row count of a parquet file before querying.
    \item \texttt{query\_parquet\_files}: Executes SQL queries over parquet files (supports SELECT, WHERE, JOIN, aggregations).
    \item \texttt{think\_tool}: A reflection tool that records the agent's internal reasoning. The prompt instructs the agent to call this tool after each query to summarize findings and plan the next investigation step.
\end{itemize}

\noindent The prompt recommends a tool-call budget of $\sim$50 calls per case and enforces a hard cap of 100 at the runtime layer; the LangGraph recursion limit is set well above this so the cap is the binding constraint. All models share the same tool definitions and identical API retry logic (exponential backoff with max 5 retries).

\subsection{Output Schemas: Ground Truth and Agent Answer}
\label{sec:appendix_causalgraph}

Evaluation uses two distinct schemas: the released ground-truth graph in each case directory, and the structured JSON the agent emits.
The ground-truth schema preserves per-node states and onset timestamps for process-level metrics; the agent schema is a smaller, verifiable contract built from \texttt{(service, fault\_kind)} root-cause claims, propagation edges, and SQL evidence.

\textbf{Ground-truth representation (released artifact).}
Each case directory ships a \texttt{causal\_graph.json} file in the \texttt{CausalGraph} schema below.
This is the per-case process-level label produced by the \frameworkName{} pipeline (Section~\ref{sec:methodology}).
Every node carries a controlled-vocabulary state and an onset timestamp, so the process-level metrics in Section~\ref{sec:appendix_metrics} can be computed directly from this representation.

\begin{schemabox}[\textsc{Schema}~~Ground-truth CausalGraph JSON]
{
  "nodes": [
    {"component": "ts-order-service",
     "state": ["HIGH_ERROR_RATE"],
     "timestamp": 1718234567000000000},
    {"component": "ts-travel-service",
     "state": ["TIMEOUT"],
     "timestamp": 1718234569000000000},
    {"component": "ts-station-service",
     "state": ["UNAVAILABLE"],
     "timestamp": 1718234565000000000}
  ],
  "edges": [
    {"source": "ts-station-service",
     "target": "ts-travel-service"},
    {"source": "ts-travel-service",
     "target": "ts-order-service"}
  ],
  "root_causes": [
    {"component": "ts-station-service",
     "state": ["UNAVAILABLE"],
     "timestamp": 1718234565000000000}
  ],
  "alarm_nodes": [
    {"component": "span|loadgenerator::POST /api/order"}
  ]
}
\end{schemabox}

\noindent The four top-level fields are:
\begin{itemize}[nosep, leftmargin=1.5em]
    \item \textbf{\texttt{nodes}}: all involved components, each tagged with abnormal state(s) and an onset timestamp.
    \item \textbf{\texttt{edges}}: directed causal arcs; an edge $(\texttt{source},\texttt{target})$ means the failure of \texttt{source} caused the failure of \texttt{target}.
    \item \textbf{\texttt{root\_causes}}: the subset of \texttt{nodes} that initiated the cascade.
    \item \textbf{\texttt{alarm\_nodes}}: the downstream nodes observed to behave abnormally during the injection window, used as targets for path-reachability scoring.
\end{itemize}
\noindent In the example, \texttt{ts-station-service} became \texttt{UNAVAILABLE} at $t_0$, propagating to \texttt{ts-travel-service} (\texttt{TIMEOUT}) and then \texttt{ts-order-service} (\texttt{HIGH\_ERROR\_RATE}), yielding a three-hop path.

\textbf{Agent answer schema.}
The agent emits a strict-JSON \texttt{AgentRCAOutput} with two flat lists.
\texttt{root\_causes} contains entries of the form (\texttt{service}, \texttt{fault\_kind}, \texttt{evidence[]}), where \texttt{fault\_kind} is drawn from a controlled enum and \texttt{evidence} is non-empty.
\texttt{propagation} contains directed (\texttt{from}, \texttt{to}) edges, each also with non-empty \texttt{evidence}.
The schema is uniform across single-fault and hybrid cases; the agent fills \texttt{root\_causes} with as many entries as the evidence supports.

\begin{schemabox}[\textsc{Schema}~~Agent answer (\texttt{AgentRCAOutput})]
{
  "root_causes": [
    {"service": "ts-station-service",
     "fault_kind": "pod_unavailable",
     "evidence": [
       {"kind": "metric",
        "sql": "SELECT time, value FROM abnormal_metrics WHERE service_name = 'ts-station-service' AND metric LIKE 'pod_phase' ORDER BY time",
        "claim": "ts-station-service pod is in NotReady throughout the window"}
     ]}
  ],
  "propagation": [
    {"from": "ts-station-service", "to": "ts-travel-service",
     "evidence": [
       {"kind": "trace",
        "sql": "SELECT span_name, attr.status_code, COUNT(*) FROM abnormal_traces WHERE service_name = 'ts-travel-service' AND span_name LIKE '%station%' GROUP BY 1,2",
        "claim": "ts-travel-service calls into ts-station-service return TIMEOUT"}
     ]}
  ]
}
\end{schemabox}

\noindent Each \texttt{Evidence} item declares (i)~its modality (\texttt{kind}~$\in$~\{\texttt{metric},\texttt{trace},\texttt{log}\}), (ii)~the actual DuckDB \texttt{sql} it claims is supported by the case parquets, and (iii)~a $\leq$20-word \texttt{claim} stating what the rows demonstrate.
The contract requires every \texttt{root\_cause} and every \texttt{propagation} edge to carry at least one evidence item, and the SQL must be re-executable by the judge against the case directory.
Service names are normalized at match time (lowercase, hyphens and underscores stripped) to absorb naming-style drift.

Both schemas feed the evaluation pipeline: the \texttt{(service, fault\_kind)} multiset and the propagation edge set are extracted from the agent answer, while their counterparts plus the alarm-node set are read off the ground-truth graph.
Section~\ref{sec:appendix_metrics} defines the per-axis comparisons.

\subsection{Prompt Design}
\label{sec:appendix_prompt}

The agent runs a two-phase prompt.
The \emph{investigation phase} (\texttt{RCA\_ANALYSIS\_SP/UP}) lets the model freely query the case parquets via tool calls and fixes the goal and playbook without prescribing the output JSON.
The \emph{synthesis phase} (\texttt{COMPRESS\_FINDINGS\_SP/UP}) converts the accumulated tool messages into one strict-JSON \texttt{AgentRCAOutput} (Section~\ref{sec:appendix_causalgraph}) using a canonical contract block spliced in from the SDK at format time.

\subsubsection{Investigation Phase}

\begin{systempromptbox}[\textsc{System}: RCA\_ANALYSIS\_SP]
You are a Root Cause Analysis (RCA) expert investigating a microservices incident.
For context, today's date is \{date\}.

\#\# Goal

Identify the root cause(s) of the SLO violation from the telemetry. There may be more than one. Let the data tell you how many faults there are and where they sit.

Your job in this phase is INVESTIGATION ONLY. A separate synthesis step will turn your findings into the final structured output. Do not try to emit JSON or final answers here.

\#\# Available data

Parquets in this case directory:
$\bullet$ abnormal\_metrics.parquet, abnormal\_traces.parquet, abnormal\_logs.parquet
$\bullet$ abnormal\_metrics\_histogram.parquet, abnormal\_metrics\_sum.parquet
$\bullet$ normal\_metrics.parquet, normal\_traces.parquet, normal\_logs.parquet

Common columns:
$\bullet$ \textbf{metrics}: time, metric, value, service\_name, attr.k8s.pod.name, attr.k8s.namespace.name
$\bullet$ \textbf{metrics\_sum}: same shape; carries jvm.cpu.*, jvm.memory.*, jvm.thread.count, jvm.gc.*, container.cpu.*, container.memory.*
$\bullet$ \textbf{metrics\_histogram}: time, metric, service\_name, count, sum, min, max, attr.jvm.gc.action, attr.jvm.gc.name
$\bullet$ \textbf{traces}: time, trace\_id, span\_id, parent\_span\_id, span\_name, service\_name, duration, attr.status\_code, attr.http.request.method, attr.http.response.status\_code
$\bullet$ \textbf{logs}: time, trace\_id, span\_id, level, service\_name, message

\#\# Tools

1. \texttt{query\_parquet\_files}: DuckDB SQL on parquets in this case dir.
2. \texttt{list\_tables\_in\_directory}: list parquets.
3. \texttt{get\_schema}: column types of a parquet.
4. \texttt{think\_tool}: REQUIRED after each query; summarize, plan next step.

\#\# Hard limits

$\bullet$ Tool-call budget: aim for $\sim$50 calls; extend if the evidence genuinely warrants it. Hard cap is 100, at which point the runtime forces a stop.
$\bullet$ Spend the budget efficiently: \texttt{list\_tables\_in\_directory} once, \texttt{get\_schema} on the files you actually plan to query, then spend the rest on \texttt{query\_parquet\_files}.

\#\# Investigation playbook

1. \texttt{list\_tables\_in\_directory} to confirm the parquet files.
2. \texttt{get\_schema} on the relevant ones (start with \texttt{abnormal\_traces}).
3. Diff abnormal vs normal: error rates, latency, status codes, log levels.
4. Trace the call chain (\texttt{parent\_span\_id} $\to$ \texttt{span\_id}) to find the earliest service whose own work, not its dependency's, went wrong.
5. Decide every root cause and every propagation edge. More than one root cause is possible; note each separately when evidence supports it.
\end{systempromptbox}

\begin{userpromptbox}[\textsc{User}: RCA\_ANALYSIS\_UP]
\{incident\_description\}
\end{userpromptbox}

\subsubsection{Synthesis Phase}

\begin{systempromptbox}[\textsc{System}: COMPRESS\_FINDINGS\_SP]
You are an RCA synthesizer. Today's date is \{date\}.

Your job: convert the investigation messages above into a single STRICT JSON object matching the RCA schema. Output JSON only, with no markdown, commentary, or prose.

\{agent\_contract\} \quad (spliced canonical block; see ``Agent Contract'' below.)

\#\# Filtering investigation messages
$\bullet$ Include: query results showing anomalies, errors, or elevated latency.
$\bullet$ Exclude: \texttt{think\_tool} outputs (the agent's own reasoning notes).
\end{systempromptbox}

\begin{userpromptbox}[\textsc{User}: COMPRESS\_FINDINGS\_UP]
\#\# Incident
\{incident\_description\}

Synthesize the investigation messages above into ONE JSON object as specified in the system prompt. No prefix, no suffix, no markdown fences, no explanation. Begin output now.
\end{userpromptbox}

\subsubsection{Agent Contract (spliced into synthesis)}

The synthesis prompt does not hand-maintain the output schema or the fault-kind enumeration.
Both are pulled at agent-startup from the SDK, so the contract version follows the SDK version automatically.
The block below is the canonical text spliced in via \verb|{agent_contract}|; it states the JSON shape of \texttt{AgentRCAOutput} (Section~\ref{sec:appendix_causalgraph}), enumerates the \texttt{fault\_kind} vocabulary, and lists the field, SQL, and hard rules.

\begin{systempromptbox}[\textsc{Spliced}: agent\_contract]
\#\# Output schema (STRICT JSON, no markdown / commentary)

JSON shape of \texttt{AgentRCAOutput} as in Section~\ref{sec:appendix_causalgraph}: \texttt{root\_causes[]} of \texttt{\{service, fault\_kind, evidence[]\}}, and \texttt{propagation[]} of \texttt{\{from, to, evidence[]\}}, where every \texttt{evidence} is \texttt{\{kind, sql, claim\}}.

\#\# fault\_kind enum (pick exactly one per root\_cause)

pod\_failure, pod\_unavailable, network\_delay, network\_loss, network\_partition, network\_corrupt, network\_duplicate, network\_bandwidth\_limit, http\_aborted, http\_slow, http\_payload\_modified, http\_response\_status\_modified, cpu\_stress, jvm\_thread\_cpu\_stress, mem\_stress, jvm\_heap\_stress, jvm\_gc\_pressure, jvm\_method\_exception, jvm\_jdbc\_exception, jvm\_method\_latency, jvm\_jdbc\_latency, jvm\_method\_mutated, dns\_resolution\_failed, dns\_resolution\_wrong, clock\_skew

\#\# Field rules
$\bullet$ Service names must match strings present in the data; do not invent.
$\bullet$ One entry per distinct root cause; do NOT collapse multiple distinct faults into a single \texttt{root\_cause}.

\#\# SQL rules
$\bullet$ DuckDB on this case dir's parquets.
$\bullet$ SELECT only; one statement per \texttt{sql} field. No DDL/DML/ATTACH/etc.
$\bullet$ Reference parquets via \texttt{read\_parquet('<file>.parquet')} with bare filenames, OR directly \texttt{FROM abnormal\_traces} (each \texttt{*.parquet} is mounted as a same-named view).
$\bullet$ Each \texttt{evidence.sql} MUST be the actual SQL, not a description.
$\bullet$ Each \texttt{evidence.claim} is the $\leq$20-word natural-language assertion the SQL rows back.

\#\# Hard rules
$\bullet$ EVERY \texttt{root\_cause} carries $\geq$1 \texttt{evidence} (real runnable SQL).
$\bullet$ EVERY propagation edge carries $\geq$1 \texttt{evidence}. Do not emit edges you cannot back with a concrete trace or metric query.
\end{systempromptbox}

\subsubsection{Key Design Choices}

\begin{itemize}[nosep, leftmargin=1.5em]
    \item \textbf{Investigation/synthesis separation.} The investigation prompt withholds the JSON schema so the model spends tokens on collecting evidence rather than packing partial answers. The synthesis prompt then receives the canonical contract block alongside the full tool history, eliminating schema drift between agent answer and judge.
    \item \textbf{Controlled fault-kind vocabulary.} The \texttt{fault\_kind} enum is observable-effect-based: two chaos primitives whose telemetry is indistinguishable share a kind, and two whose signals are distinguishable get separate kinds. This lets the matcher score classification (\emph{fault\_kind\_accuracy}) deterministically without an LLM-as-judge.
    \item \textbf{Mandatory SQL evidence.} Every claim, including both root causes and propagation edges, must cite at least one DuckDB-runnable SQL. The judge re-executes each SQL on the case directory; un-runnable evidence (\emph{sql\_executable\_rate}) and rows that do not back the claim (\emph{evidence\_support\_rate}, Section~\ref{sec:appendix_metrics}) each surface as a scoreable axis.
    \item \textbf{No topology leakage.} The agent is \emph{not} given the service dependency graph; service relations must be inferred from trace data, testing investigative ability rather than pattern matching over a known graph.
    \item \textbf{Paired normal/abnormal telemetry.} Both pre- and post-injection parquets ship in the case directory, so the agent can run differential queries (error rates, latency tails, log volumes) on identical schemas.
\end{itemize}

\subsection{Model Configurations}
\label{sec:appendix_model_config}

We evaluate Claude Opus 4.7, Claude Sonnet 4.6, Gemini 3.1 Pro, MiMo 2.5 Pro, GLM 5.1, Kimi K2.6, DeepSeek V4 Pro, Qwen3.6-Max, Hy 3.0 Preview, Seed 2.0 Pro, and MiniMax M2.7 in the main results, plus GPT-5.4 and GPT-5.5 reported separately in Appendix~\ref{sec:appendix_broken_coverage}.

All models use identical system prompts, tool definitions, and temperature (0.7) for reproducibility.
GPT-5.4 and GPT-5.5 have limited benchmark coverage and are reported separately in Appendix~\ref{sec:appendix_broken_coverage}.

\section{Evaluation Protocol and Metric Definitions}
\label{sec:appendix_metrics}

We organize our evaluation metrics into two layers: \textit{outcome} (set classification of the predicted root-cause set against ground truth) and \textit{process} (causal reasoning quality on the predicted graph plus evidence executability), complemented by \textit{dataset difficulty} signals that characterize each instance.

\textbf{Claim-to-Metric Mapping.}
Each headline claim in Section~\ref{sec:exp_benchmark} is anchored to a specific metric defined below:
\begin{itemize}[nosep, leftmargin=1.5em]
    \item \emph{Outcome scores understate the failure; the lenient process check exposes ungrounded diagnoses} (Finding~1): measured by the share of cases satisfying AnySvc but not PR, and by the conditional rate $\text{PR}/\text{AnySvc}$ in Table~\ref{tab:llm_evaluation}.
    \item \emph{Edge F1 lags Node F1 universally} (Finding~2): measured by the per-row Node F1 vs.\ Edge F1 columns.
    \item \emph{No single best model} (Finding~3): measured by the distribution of per-column extrema (bolded in Table~\ref{tab:llm_evaluation}) across distinct models.
    \item \emph{Per-system difficulty is uneven} (Appendix Section~\ref{sec:appendix_per_system}): measured by the spread of EM and PR across the TrainTicket / OTel Demo / Hotel Reservation rows of Appendix Table~\ref{tab:llm_evaluation_per_system}.
\end{itemize}

\subsection{Outcome-Level Metrics}
\label{sec:appendix_outcome_metrics}

The outcome-level metrics grade only the agent's predicted root-cause set against the ground-truth root-cause set, ignoring the predicted causal graph.
Let $\hat{R}_q$ denote the set of (service, fault\_kind) pairs the agent emits as root causes for case $q$, $R^*_q$ the corresponding ground-truth set, and $\mathcal{Q}$ the set of all evaluation queries.
Service names are normalized as in Section~\ref{sec:appendix_process_metrics} before comparison; for network-class faults, both endpoints of the netem rule are accepted because the rule is observable on either side.

\textbf{AnySvc (any service hit, kind-agnostic).}
\begin{equation}
    \text{AnySvc} = \frac{1}{|\mathcal{Q}|} \sum_{q \in \mathcal{Q}} \mathbb{1}\!\left[\,\exists\,(\hat{s}, \hat{k}) \in \hat{R}_q,\ (s^*, k^*) \in R^*_q : \hat{s} = s^*\,\right]
\end{equation}
AnySvc is satisfied when at least one predicted root-cause service appears in the ground truth, irrespective of the predicted fault kind.

\textbf{Precision, Recall, F1 on (service, fault\_kind) pairs.}
For each case $q$, the per-case precision and recall on the pair set are
\begin{align}
    \text{P}_q = \frac{|\hat{R}_q \cap R^*_q|}{|\hat{R}_q|},\quad
    \text{R}_q = \frac{|\hat{R}_q \cap R^*_q|}{|R^*_q|},\quad
    \text{F1}_q = \frac{2\,\text{P}_q\,\text{R}_q}{\text{P}_q + \text{R}_q}
\end{align}
where intersection uses HIT-and-equivalence semantics (a predicted pair counts as matched if there is a ground-truth pair with the same normalized service and an equivalent fault\_kind in the equivalence vocabulary of Section~\ref{sec:appendix_kind_vocab}).
We report the mean of $\text{P}_q$, $\text{R}_q$, and $\text{F1}_q$ over $\mathcal{Q}$.
Recall captures fractional coverage of the ground-truth pairs and is therefore a strictly finer measurement than AnySvc, which only checks whether at least one service is named.

\textbf{Exact Match (EM).}
\begin{equation}
    \text{EM} = \frac{1}{|\mathcal{Q}|} \sum_{q \in \mathcal{Q}} \mathbb{1}\!\left[\,\hat{R}_q = R^*_q\,\right]
\end{equation}
EM requires the predicted root-cause set to coincide with the ground truth as sets, penalizing both missing and spurious entries.

\subsection{Process-Level Metrics}
\label{sec:appendix_process_metrics}

These metrics evaluate the quality of the agent's causal reasoning by comparing the predicted CausalGraph against the verified causal subgraph $\mathcal{G}^*$.
We decompose the evaluation into \textit{primary} metrics (service-level, deterministic matching) and \textit{secondary} metrics (component-level, with LLM-assisted semantic matching for name normalization).

\textbf{Loadgen filtering.}
Edges incident on the synthetic load-generator endpoint are removed from both $\hat{\mathcal{E}}_q$ and $\mathcal{E}^*_q$ before any graph-shape metric (Edge F1, Node F1, Path Reachability) is computed.
Loadgen is an evaluation-time client and not part of the system under test, so neither admitting nor omitting loadgen-touching edges should be scored.

\subsubsection{Primary Metrics (Service-Level)}

All primary metrics operate on service-level representations. 
Service names are normalized (lowercased, \texttt{ts-} prefix stripped, hyphens removed) before comparison.

\textbf{Edge F1.}
Let $\hat{\mathcal{E}}$ and $\mathcal{E}^*$ be the service-level edge sets of the predicted and ground-truth causal graphs, respectively:
\begin{align}
    \text{Edge Precision} &= \frac{|\hat{\mathcal{E}} \cap \mathcal{E}^*|}{|\hat{\mathcal{E}}|}, \quad
    \text{Edge Recall} = \frac{|\hat{\mathcal{E}} \cap \mathcal{E}^*|}{|\mathcal{E}^*|}
\end{align}
Edge F1 is the harmonic mean.

\textbf{Node F1.}
Let $\hat{\mathcal{V}}$ and $\mathcal{V}^*$ be the service-level node sets:
\begin{align}
    \text{Node Precision} &= \frac{|\hat{\mathcal{V}} \cap \mathcal{V}^*|}{|\hat{\mathcal{V}}|}, \quad
    \text{Node Recall} = \frac{|\hat{\mathcal{V}} \cap \mathcal{V}^*|}{|\mathcal{V}^*|}
\end{align}
Node F1 measures whether the agent correctly identified the set of affected services.

\textbf{Boundary case.} When both the predicted and ground-truth sets are empty for a metric (e.g., both graphs have zero edges), we assign a perfect score of 1.0, since the agent correctly predicted the absence of that element.

\subsubsection{SQL Exec}
\label{sec:appendix_sql_exec}

The agent answer schema (Section~\ref{sec:appendix_causalgraph}) ties every claim to at least one piece of SQL evidence.
Every \texttt{evidence.sql} in the agent answer is re-executed on a fresh in-memory DuckDB connection with each \texttt{*.parquet} of the case directory mounted as a same-named view.
The result is one of three statuses: \texttt{OK} (executed, $\geq 1$ row), \texttt{EMPTY} (executed, 0 rows), or \texttt{SQL\_ERROR} (parse or runtime failure).
Per case,
\begin{equation}
    \text{SQL Exec} = \frac{|\{e \in \mathcal{E}_{\text{ev},q}: \text{status}(e) = \texttt{OK}\}|}{|\mathcal{E}_{\text{ev},q}|}
\end{equation}
where $\mathcal{E}_{\text{ev},q}$ is the union of all evidence items in $q$'s root\_causes and propagation edges.
This axis is mechanical; whether the rows actually back the claim is not scored here.

\subsubsection{Diagnostic Reasoning Axes}

\textbf{Path Reachability (PR).}
Path reachability measures the fraction of all instances where the agent both names a correct root-cause service \emph{and} constructs at least one valid directed path from that service to a ground-truth alarm node in its predicted graph:
\begin{equation}
    \text{PR} = \frac{1}{|\mathcal{Q}|} \sum_{q \in \mathcal{Q}} \mathbb{1}\left[\text{AnySvc}_q = 1 \;\wedge\; \exists\, \text{path } \hat{r}_q \leadsto \hat{s}_q \text{ in } \hat{\mathcal{G}}_q \mid \hat{r}_q \in R^*_q,\, \hat{s}_q \in A^*_q \right]
\end{equation}
where $R^*_q$ and $A^*_q$ are the ground-truth root cause and alarm-node sets for instance $q$, and $\text{AnySvc}_q = 1$ when at least one of the agent's predicted root-cause services lies in $R^*_q$ (kind-agnostic).
Anchoring on AnySvc rather than on a kind-correct hit lets PR vary independently from the fault-kind-tagged Recall: PR isolates whether the agent constructed a faithful chain, while Recall captures whether it also labelled the fault kind correctly.
Cases where the agent named no correct service receive PR${} = 0$ by definition, since no service-correct anchor exists for the path.
PR is therefore strictly upper-bounded by AnySvc, and the share of AnySvc cases failing PR quantifies how often a correctly localized service lacks even a single actionable propagation path.

\paragraph{HIT-anchored variant.}
For backward comparison we also report a HIT-anchored variant of PR (Appendix Table~\ref{tab:llm_evaluation_pr_hit}), which restricts the anchor to root-cause claims that match both service \emph{and} fault kind.
The HIT-anchored variant is strictly upper-bounded by Recall and is what the SDK stores by default; the AnySvc-anchored variant in the main table is computed locally from the same trajectories.

\section{Auxiliary Evaluation Tables}
\label{sec:appendix_aux_tables}

\subsection{Per-system breakdown}
\label{sec:appendix_per_system}

Table~\ref{tab:llm_evaluation_per_system} expands the pooled main-paper table (Table~\ref{tab:llm_evaluation}) into per-system rows for TrainTicket, OTel Demo, and Hotel Reservation.
The pooled \textbf{All} row of each model matches the corresponding row in the main table.

The per-system spread is wide enough to dominate the cross-model spread on the easier metrics.
On Hotel Reservation (\evalCountHr{} cases, 9 services, mean longest path \meanLpHr{} hops) Opus reaches \opusLiteHrAnySvc{}\% AnySvc and \opusLiteHrEM{}\% EM, while on TrainTicket (\evalCountTt{} cases, 44 services, mean longest path \meanLpTt{} hops) the same model drops to \opusLiteTtAnySvc{}\% AnySvc and \opusLiteTtEM{}\% EM; OpenTelemetry Demo (\evalCountOtel{} cases, 15 services, \meanLpOtel{} hops) sits between, at \opusLiteOtelAnySvc{}\% AnySvc and \opusLiteOtelEM{}\% EM.
Both service count and chain depth track the difficulty ordering in the same direction.
Finding~(2) in Section~\ref{sec:exp_benchmark} suggests path depth is the load-bearing factor: Edge F1 trails Node F1 by $\sim$19\,pp pooled, and the deficit compounds along longer paths.
Pooled scores across systems with disparate chain depth are therefore sensitive to system-mix manipulation; the per-system rows below are the direct read for any narrow-regime comparison.

\begin{table}[t]
\centering
\caption{Per-system breakdown of Table~\ref{tab:llm_evaluation} (TrainTicket, OTel Demo, Hotel Reservation; pooled aggregate in the \textbf{All} row; percentages).
}
\label{tab:llm_evaluation_per_system}
\resizebox{0.9\textwidth}{!}{%
\begin{tabular}{ll ccccc cccc}
\toprule
 & & \multicolumn{5}{c}{\textit{Outcome}} & \multicolumn{4}{c}{\textit{Process}} \\
\cmidrule(lr){3-7} \cmidrule(lr){8-11}
Model & System & EM$\uparrow$ & F1$\uparrow$ & Precision$\uparrow$ & Recall$\uparrow$ & AnySvc$\uparrow$ & PR$\uparrow$ & Node F1$\uparrow$ & Edge F1$\uparrow$ & SQL Exec$\uparrow$ \\
\midrule
\multirow{4}{*}{Claude Opus 4.7}
 & TrainTicket  & \opusLiteTtEM & \opusLiteTtF & \opusLiteTtPrecision & \opusLiteTtRecall & \opusLiteTtAnySvc & \opusLiteTtPathReach & \opusLiteTtNodeF & \opusLiteTtEdgeF & \opusLiteTtSqlExec \\
 & OTel Demo  & \opusLiteOtelEM & \opusLiteOtelF & \opusLiteOtelPrecision & \opusLiteOtelRecall & \opusLiteOtelAnySvc & \opusLiteOtelPathReach & \opusLiteOtelNodeF & \opusLiteOtelEdgeF & \opusLiteOtelSqlExec \\
 & Hotel Res.\  & \opusLiteHrEM & \opusLiteHrF & \opusLiteHrPrecision & \opusLiteHrRecall & \opusLiteHrAnySvc & \opusLiteHrPathReach & \opusLiteHrNodeF & \opusLiteHrEdgeF & \opusLiteHrSqlExec \\
 & \textbf{All} & \textbf{\opusLiteEM} & \textbf{\opusLiteF} & \textbf{\opusLitePrecision} & \textbf{\opusLiteRecall} & \textbf{\opusLiteAnySvc} & \textbf{\opusLitePathReach} & \textbf{\opusLiteNodeF} & \textbf{\opusLiteEdgeF} & \textbf{\opusLiteSqlExec} \\
\midrule
\multirow{4}{*}{Gemini 3.1 Pro}
 & TrainTicket  & \geminiLiteTtEM & \geminiLiteTtF & \geminiLiteTtPrecision & \geminiLiteTtRecall & \geminiLiteTtAnySvc & \geminiLiteTtPathReach & \geminiLiteTtNodeF & \geminiLiteTtEdgeF & \geminiLiteTtSqlExec \\
 & OTel Demo  & \geminiLiteOtelEM & \geminiLiteOtelF & \geminiLiteOtelPrecision & \geminiLiteOtelRecall & \geminiLiteOtelAnySvc & \geminiLiteOtelPathReach & \geminiLiteOtelNodeF & \geminiLiteOtelEdgeF & \geminiLiteOtelSqlExec \\
 & Hotel Res.\  & \geminiLiteHrEM & \geminiLiteHrF & \geminiLiteHrPrecision & \geminiLiteHrRecall & \geminiLiteHrAnySvc & \geminiLiteHrPathReach & \geminiLiteHrNodeF & \geminiLiteHrEdgeF & \geminiLiteHrSqlExec \\
 & \textbf{All} & \textbf{\geminiLiteEM} & \textbf{\geminiLiteF} & \textbf{\geminiLitePrecision} & \textbf{\geminiLiteRecall} & \textbf{\geminiLiteAnySvc} & \textbf{\geminiLitePathReach} & \textbf{\geminiLiteNodeF} & \textbf{\geminiLiteEdgeF} & \textbf{\geminiLiteSqlExec} \\
\midrule
\multirow{4}{*}{Claude Sonnet 4.6}
 & TrainTicket  & \sonnetLiteTtEM & \sonnetLiteTtF & \sonnetLiteTtPrecision & \sonnetLiteTtRecall & \sonnetLiteTtAnySvc & \sonnetLiteTtPathReach & \sonnetLiteTtNodeF & \sonnetLiteTtEdgeF & \sonnetLiteTtSqlExec \\
 & OTel Demo  & \sonnetLiteOtelEM & \sonnetLiteOtelF & \sonnetLiteOtelPrecision & \sonnetLiteOtelRecall & \sonnetLiteOtelAnySvc & \sonnetLiteOtelPathReach & \sonnetLiteOtelNodeF & \sonnetLiteOtelEdgeF & \sonnetLiteOtelSqlExec \\
 & Hotel Res.\  & \sonnetLiteHrEM & \sonnetLiteHrF & \sonnetLiteHrPrecision & \sonnetLiteHrRecall & \sonnetLiteHrAnySvc & \sonnetLiteHrPathReach & \sonnetLiteHrNodeF & \sonnetLiteHrEdgeF & \sonnetLiteHrSqlExec \\
 & \textbf{All} & \textbf{\sonnetLiteEM} & \textbf{\sonnetLiteF} & \textbf{\sonnetLitePrecision} & \textbf{\sonnetLiteRecall} & \textbf{\sonnetLiteAnySvc} & \textbf{\sonnetLitePathReach} & \textbf{\sonnetLiteNodeF} & \textbf{\sonnetLiteEdgeF} & \textbf{\sonnetLiteSqlExec} \\
\midrule
\multirow{4}{*}{MiMo 2.5 Pro}
 & TrainTicket  & \mimoLiteTtEM & \mimoLiteTtF & \mimoLiteTtPrecision & \mimoLiteTtRecall & \mimoLiteTtAnySvc & \mimoLiteTtPathReach & \mimoLiteTtNodeF & \mimoLiteTtEdgeF & \mimoLiteTtSqlExec \\
 & OTel Demo  & \mimoLiteOtelEM & \mimoLiteOtelF & \mimoLiteOtelPrecision & \mimoLiteOtelRecall & \mimoLiteOtelAnySvc & \mimoLiteOtelPathReach & \mimoLiteOtelNodeF & \mimoLiteOtelEdgeF & \mimoLiteOtelSqlExec \\
 & Hotel Res.\  & \mimoLiteHrEM & \mimoLiteHrF & \mimoLiteHrPrecision & \mimoLiteHrRecall & \mimoLiteHrAnySvc & \mimoLiteHrPathReach & \mimoLiteHrNodeF & \mimoLiteHrEdgeF & \mimoLiteHrSqlExec \\
 & \textbf{All} & \textbf{\mimoLiteEM} & \textbf{\mimoLiteF} & \textbf{\mimoLitePrecision} & \textbf{\mimoLiteRecall} & \textbf{\mimoLiteAnySvc} & \textbf{\mimoLitePathReach} & \textbf{\mimoLiteNodeF} & \textbf{\mimoLiteEdgeF} & \textbf{\mimoLiteSqlExec} \\
\midrule
\multirow{4}{*}{GLM 5.1}
 & TrainTicket  & \glmLiteTtEM & \glmLiteTtF & \glmLiteTtPrecision & \glmLiteTtRecall & \glmLiteTtAnySvc & \glmLiteTtPathReach & \glmLiteTtNodeF & \glmLiteTtEdgeF & \glmLiteTtSqlExec \\
 & OTel Demo  & \glmLiteOtelEM & \glmLiteOtelF & \glmLiteOtelPrecision & \glmLiteOtelRecall & \glmLiteOtelAnySvc & \glmLiteOtelPathReach & \glmLiteOtelNodeF & \glmLiteOtelEdgeF & \glmLiteOtelSqlExec \\
 & Hotel Res.\  & \glmLiteHrEM & \glmLiteHrF & \glmLiteHrPrecision & \glmLiteHrRecall & \glmLiteHrAnySvc & \glmLiteHrPathReach & \glmLiteHrNodeF & \glmLiteHrEdgeF & \glmLiteHrSqlExec \\
 & \textbf{All} & \textbf{\glmLiteEM} & \textbf{\glmLiteF} & \textbf{\glmLitePrecision} & \textbf{\glmLiteRecall} & \textbf{\glmLiteAnySvc} & \textbf{\glmLitePathReach} & \textbf{\glmLiteNodeF} & \textbf{\glmLiteEdgeF} & \textbf{\glmLiteSqlExec} \\
\midrule
\multirow{4}{*}{Kimi K2.6}
 & TrainTicket  & \kimiLiteTtEM & \kimiLiteTtF & \kimiLiteTtPrecision & \kimiLiteTtRecall & \kimiLiteTtAnySvc & \kimiLiteTtPathReach & \kimiLiteTtNodeF & \kimiLiteTtEdgeF & \kimiLiteTtSqlExec \\
 & OTel Demo  & \kimiLiteOtelEM & \kimiLiteOtelF & \kimiLiteOtelPrecision & \kimiLiteOtelRecall & \kimiLiteOtelAnySvc & \kimiLiteOtelPathReach & \kimiLiteOtelNodeF & \kimiLiteOtelEdgeF & \kimiLiteOtelSqlExec \\
 & Hotel Res.\  & \kimiLiteHrEM & \kimiLiteHrF & \kimiLiteHrPrecision & \kimiLiteHrRecall & \kimiLiteHrAnySvc & \kimiLiteHrPathReach & \kimiLiteHrNodeF & \kimiLiteHrEdgeF & \kimiLiteHrSqlExec \\
 & \textbf{All} & \textbf{\kimiLiteEM} & \textbf{\kimiLiteF} & \textbf{\kimiLitePrecision} & \textbf{\kimiLiteRecall} & \textbf{\kimiLiteAnySvc} & \textbf{\kimiLitePathReach} & \textbf{\kimiLiteNodeF} & \textbf{\kimiLiteEdgeF} & \textbf{\kimiLiteSqlExec} \\
\midrule
\multirow{4}{*}{DeepSeek V4 Pro}
 & TrainTicket  & \dsvfourLiteTtEM & \dsvfourLiteTtF & \dsvfourLiteTtPrecision & \dsvfourLiteTtRecall & \dsvfourLiteTtAnySvc & \dsvfourLiteTtPathReach & \dsvfourLiteTtNodeF & \dsvfourLiteTtEdgeF & \dsvfourLiteTtSqlExec \\
 & OTel Demo  & \dsvfourLiteOtelEM & \dsvfourLiteOtelF & \dsvfourLiteOtelPrecision & \dsvfourLiteOtelRecall & \dsvfourLiteOtelAnySvc & \dsvfourLiteOtelPathReach & \dsvfourLiteOtelNodeF & \dsvfourLiteOtelEdgeF & \dsvfourLiteOtelSqlExec \\
 & Hotel Res.\  & \dsvfourLiteHrEM & \dsvfourLiteHrF & \dsvfourLiteHrPrecision & \dsvfourLiteHrRecall & \dsvfourLiteHrAnySvc & \dsvfourLiteHrPathReach & \dsvfourLiteHrNodeF & \dsvfourLiteHrEdgeF & \dsvfourLiteHrSqlExec \\
 & \textbf{All} & \textbf{\dsvfourLiteEM} & \textbf{\dsvfourLiteF} & \textbf{\dsvfourLitePrecision} & \textbf{\dsvfourLiteRecall} & \textbf{\dsvfourLiteAnySvc} & \textbf{\dsvfourLitePathReach} & \textbf{\dsvfourLiteNodeF} & \textbf{\dsvfourLiteEdgeF} & \textbf{\dsvfourLiteSqlExec} \\
\midrule
\multirow{4}{*}{Qwen3.6-Max}
 & TrainTicket  & \qwenLiteTtEM & \qwenLiteTtF & \qwenLiteTtPrecision & \qwenLiteTtRecall & \qwenLiteTtAnySvc & \qwenLiteTtPathReach & \qwenLiteTtNodeF & \qwenLiteTtEdgeF & \qwenLiteTtSqlExec \\
 & OTel Demo  & \qwenLiteOtelEM & \qwenLiteOtelF & \qwenLiteOtelPrecision & \qwenLiteOtelRecall & \qwenLiteOtelAnySvc & \qwenLiteOtelPathReach & \qwenLiteOtelNodeF & \qwenLiteOtelEdgeF & \qwenLiteOtelSqlExec \\
 & Hotel Res.\  & \qwenLiteHrEM & \qwenLiteHrF & \qwenLiteHrPrecision & \qwenLiteHrRecall & \qwenLiteHrAnySvc & \qwenLiteHrPathReach & \qwenLiteHrNodeF & \qwenLiteHrEdgeF & \qwenLiteHrSqlExec \\
 & \textbf{All} & \textbf{\qwenLiteEM} & \textbf{\qwenLiteF} & \textbf{\qwenLitePrecision} & \textbf{\qwenLiteRecall} & \textbf{\qwenLiteAnySvc} & \textbf{\qwenLitePathReach} & \textbf{\qwenLiteNodeF} & \textbf{\qwenLiteEdgeF} & \textbf{\qwenLiteSqlExec} \\
\midrule
\multirow{4}{*}{Hy 3.0 Preview}
 & TrainTicket  & \hyLiteTtEM & \hyLiteTtF & \hyLiteTtPrecision & \hyLiteTtRecall & \hyLiteTtAnySvc & \hyLiteTtPathReach & \hyLiteTtNodeF & \hyLiteTtEdgeF & \hyLiteTtSqlExec \\
 & OTel Demo  & \hyLiteOtelEM & \hyLiteOtelF & \hyLiteOtelPrecision & \hyLiteOtelRecall & \hyLiteOtelAnySvc & \hyLiteOtelPathReach & \hyLiteOtelNodeF & \hyLiteOtelEdgeF & \hyLiteOtelSqlExec \\
 & Hotel Res.\  & \hyLiteHrEM & \hyLiteHrF & \hyLiteHrPrecision & \hyLiteHrRecall & \hyLiteHrAnySvc & \hyLiteHrPathReach & \hyLiteHrNodeF & \hyLiteHrEdgeF & \hyLiteHrSqlExec \\
 & \textbf{All} & \textbf{\hyLiteEM} & \textbf{\hyLiteF} & \textbf{\hyLitePrecision} & \textbf{\hyLiteRecall} & \textbf{\hyLiteAnySvc} & \textbf{\hyLitePathReach} & \textbf{\hyLiteNodeF} & \textbf{\hyLiteEdgeF} & \textbf{\hyLiteSqlExec} \\
\midrule
\multirow{4}{*}{Seed 2.0 Pro}
 & TrainTicket  & \seedLiteTtEM & \seedLiteTtF & \seedLiteTtPrecision & \seedLiteTtRecall & \seedLiteTtAnySvc & \seedLiteTtPathReach & \seedLiteTtNodeF & \seedLiteTtEdgeF & \seedLiteTtSqlExec \\
 & OTel Demo  & \seedLiteOtelEM & \seedLiteOtelF & \seedLiteOtelPrecision & \seedLiteOtelRecall & \seedLiteOtelAnySvc & \seedLiteOtelPathReach & \seedLiteOtelNodeF & \seedLiteOtelEdgeF & \seedLiteOtelSqlExec \\
 & Hotel Res.\  & \seedLiteHrEM & \seedLiteHrF & \seedLiteHrPrecision & \seedLiteHrRecall & \seedLiteHrAnySvc & \seedLiteHrPathReach & \seedLiteHrNodeF & \seedLiteHrEdgeF & \seedLiteHrSqlExec \\
 & \textbf{All} & \textbf{\seedLiteEM} & \textbf{\seedLiteF} & \textbf{\seedLitePrecision} & \textbf{\seedLiteRecall} & \textbf{\seedLiteAnySvc} & \textbf{\seedLitePathReach} & \textbf{\seedLiteNodeF} & \textbf{\seedLiteEdgeF} & \textbf{\seedLiteSqlExec} \\
\midrule
\multirow{4}{*}{MiniMax M2.7}
 & TrainTicket  & \minimaxLiteTtEM & \minimaxLiteTtF & \minimaxLiteTtPrecision & \minimaxLiteTtRecall & \minimaxLiteTtAnySvc & \minimaxLiteTtPathReach & \minimaxLiteTtNodeF & \minimaxLiteTtEdgeF & \minimaxLiteTtSqlExec \\
 & OTel Demo  & \minimaxLiteOtelEM & \minimaxLiteOtelF & \minimaxLiteOtelPrecision & \minimaxLiteOtelRecall & \minimaxLiteOtelAnySvc & \minimaxLiteOtelPathReach & \minimaxLiteOtelNodeF & \minimaxLiteOtelEdgeF & \minimaxLiteOtelSqlExec \\
 & Hotel Res.\  & \minimaxLiteHrEM & \minimaxLiteHrF & \minimaxLiteHrPrecision & \minimaxLiteHrRecall & \minimaxLiteHrAnySvc & \minimaxLiteHrPathReach & \minimaxLiteHrNodeF & \minimaxLiteHrEdgeF & \minimaxLiteHrSqlExec \\
 & \textbf{All} & \textbf{\minimaxLiteEM} & \textbf{\minimaxLiteF} & \textbf{\minimaxLitePrecision} & \textbf{\minimaxLiteRecall} & \textbf{\minimaxLiteAnySvc} & \textbf{\minimaxLitePathReach} & \textbf{\minimaxLiteNodeF} & \textbf{\minimaxLiteEdgeF} & \textbf{\minimaxLiteSqlExec} \\
\bottomrule
\end{tabular}%
}
\end{table}

\subsection{HIT-anchored Path Reachability}
\label{sec:appendix_pr_hit}

Table~\ref{tab:llm_evaluation_pr_hit} reports PR under the stricter HIT anchor (the SDK default): the agent must match both service \emph{and} fault kind on the anchoring root-cause claim before the path is checked.
The HIT-anchored variant is what the per-case \texttt{result.json} stores; the main-table PR (Table~\ref{tab:llm_evaluation}) relaxes the anchor to AnySvc so PR varies independently from fault-kind labelling.

\begin{table}[t]
\centering
\caption{HIT-anchored Path Reachability (PR$^{\text{HIT}}$) per (model, system).
Strictly upper-bounded by Recall. Cf.\ Table~\ref{tab:llm_evaluation} for the AnySvc-anchored PR used in the main table.}
\label{tab:llm_evaluation_pr_hit}
\resizebox{0.6\columnwidth}{!}{%
\begin{tabular}{l cccc}
\toprule
Model & TrainTicket & OTel Demo & Hotel Res.\ & \textbf{All} \\
\midrule
Claude Opus 4.7         & \opusLiteTtPathReachHit    & \opusLiteOtelPathReachHit    & \opusLiteHrPathReachHit    & \textbf{\opusLitePathReachHit}    \\
Gemini 3.1 Pro              & \geminiLiteTtPathReachHit  & \geminiLiteOtelPathReachHit  & \geminiLiteHrPathReachHit  & \textbf{\geminiLitePathReachHit}  \\
Claude Sonnet 4.6           & \sonnetLiteTtPathReachHit  & \sonnetLiteOtelPathReachHit  & \sonnetLiteHrPathReachHit  & \textbf{\sonnetLitePathReachHit}  \\
MiMo 2.5 Pro            & \mimoLiteTtPathReachHit    & \mimoLiteOtelPathReachHit    & \mimoLiteHrPathReachHit    & \textbf{\mimoLitePathReachHit}    \\
GLM 5.1                 & \glmLiteTtPathReachHit     & \glmLiteOtelPathReachHit     & \glmLiteHrPathReachHit     & \textbf{\glmLitePathReachHit}     \\
Kimi K2.6                   & \kimiLiteTtPathReachHit    & \kimiLiteOtelPathReachHit    & \kimiLiteHrPathReachHit    & \textbf{\kimiLitePathReachHit}    \\
DeepSeek V4 Pro         & \dsvfourLiteTtPathReachHit & \dsvfourLiteOtelPathReachHit & \dsvfourLiteHrPathReachHit & \textbf{\dsvfourLitePathReachHit} \\
Qwen3.6-Max             & \qwenLiteTtPathReachHit    & \qwenLiteOtelPathReachHit    & \qwenLiteHrPathReachHit    & \textbf{\qwenLitePathReachHit}    \\
Hy 3.0 Preview          & \hyLiteTtPathReachHit      & \hyLiteOtelPathReachHit      & \hyLiteHrPathReachHit      & \textbf{\hyLitePathReachHit}      \\
Seed 2.0 Pro            & \seedLiteTtPathReachHit    & \seedLiteOtelPathReachHit    & \seedLiteHrPathReachHit    & \textbf{\seedLitePathReachHit}    \\
MiniMax M2.7            & \minimaxLiteTtPathReachHit & \minimaxLiteOtelPathReachHit & \minimaxLiteHrPathReachHit & \textbf{\minimaxLitePathReachHit} \\
\bottomrule
\end{tabular}%
}
\end{table}

\subsection{Limited-coverage models}
\label{sec:appendix_broken_coverage}

GPT-5.4 and GPT-5.5 hit API budget limits before completing the full benchmark, so their sample sizes are smaller than the main-table models.
We include their partial results below for reference only; they should not be compared directly to Table~\ref{tab:llm_evaluation}.

\begin{table}[t]
\centering
\caption{GPT models with partial coverage due to budget limits (pooled across systems; for reference only).}
\label{tab:llm_evaluation_broken}
\resizebox{0.7\columnwidth}{!}{%
\begin{tabular}{l r cccc cc}
\toprule
Model & n & EM$\uparrow$ & F1$\uparrow$ & AnySvc$\uparrow$ & PR$\uparrow$ & Node F1$\uparrow$ & Edge F1$\uparrow$ \\
\midrule
GPT-5.4 & \gptFiveFourLiteN & \gptFiveFourLiteEM & \gptFiveFourLiteF & \gptFiveFourLiteAnySvc & \gptFiveFourLitePathReach & \gptFiveFourLiteNodeF & \gptFiveFourLiteEdgeF \\
GPT-5.5 & \gptFiveFiveLiteN & \gptFiveFiveLiteEM & \gptFiveFiveLiteF & \gptFiveFiveLiteAnySvc & \gptFiveFiveLitePathReach & \gptFiveFiveLiteNodeF & \gptFiveFiveLiteEdgeF \\
\bottomrule
\end{tabular}%
}
\end{table}

\newpage

\end{document}